\documentclass[preprint,12pt]{elsarticle}
\usepackage[margin=1in]{geometry}

\usepackage{times}
\usepackage{amssymb, amsmath, amsbsy}
\usepackage{makecell}  
\usepackage{graphicx}
\usepackage{float}
\usepackage{multirow}
\usepackage{xcolor}
\usepackage{soul}
\soulregister\cite7
\soulregister\ref7
\soulregister\pageref7

\usepackage{hyperref}

\begin{document}

\begin{frontmatter}



\title{A deep learning framework for solution and discovery in solid mechanics}


\author[MIT]{Ehsan Haghighat}
\address[MIT]{Massachusetts Institute of Technology, Cambridge, MA}
\author[UCB]{Maziar Raissi}
\address[UCB]{University of Colorado Boulder, Boulder, CO}
\author[PU]{Adrian Moure}
\author[PU]{Hector Gomez}
\address[PU]{Purdue University, West Lafayette, IN}
\author[MIT]{Ruben Juanes}

\begin{abstract}
We present the application of a class of deep learning, known as Physics Informed Neural Networks (PINN), to learning and discovery in solid mechanics. We explain how to incorporate the momentum balance and constitutive relations into PINN, and explore in detail the application to linear elasticity, and illustrate its extension to nonlinear problems through an example that showcases von~Mises elastoplasticity. While common PINN algorithms are based on training one deep neural network (DNN), we propose a multi-network model that results in more accurate representation of the field variables. To validate the model, we test the framework on synthetic data generated from analytical and numerical reference solutions. We study convergence of the PINN model, and show that Isogeometric Analysis (IGA) results in superior accuracy and convergence characteristics compared with classic low-order Finite Element Method (FEM). We also show the applicability of the framework for transfer learning, and find vastly accelerated convergence during network re-training. Finally, we find that honoring the physics leads to improved robustness: when trained only on a few parameters, we find that the PINN model can accurately predict the solution for a wide range of parameters new to the network---thus pointing to an important application of this framework to sensitivity analysis and surrogate modeling.
\end{abstract}




\end{frontmatter}



\section{Introduction}\label{sec:intro}
Over the past few years, there has been a revolution in the successful application of Artificial Neural Networks (ANN), also commonly referred to Deep Neural Networks (DNN) and Deep Learning (DL), in various fields including image classification, handwriting recognition, speech recognition and translation, and computer vision. These ANN approaches have led to a sea change in the performance of search engines, autonomous driving, e-commerce, and photography (see \cite{bishop2006pattern, Lecun2015, Goodfellow2016} for a review). In engineering and science, ANNs have been applied to an increasing number of areas, including geosciences \cite{Yoon2015, Bergen2019, DeVries2018, Kong2018, Ren2019}, material science \cite{Pilania2013, Butler2018, Shi2019, Brunton2019}, fluid mechanics \cite{brenner2019perspective,brunton2019machine}, genetics \cite{Libbrecht2015}, and infrastructure health monitoring \cite{Rafiei2017, Sen2019}, to name a few examples. In the solid and geomechanics community, deep learning has been used primarily for material modeling, in an attempt to replace classical constitutive models with ANNs \cite{ghaboussi1998new, kalidindi2011microstructure, Mozaffar2019}. In these applications, training of the network, i.e., evaluation of the network parameters, is carried out by minimizing the norm of the distance between the network output (prediction) and the true output (training data). In this paper, we will refer to ANNs trained in this way as ``data-driven.''

A different class of ANNs, known as Physics-Informed Neural Networks (PINN), was introduced recently \cite{rudy2019data, Raissi2019, han2018solving, bar2019learning, zhu2019physics}. This concept of ANNs was developed to endow the network model with known equations that govern the physics of a system. The training of PINNs is performed with a cost function that, in addition to data, includes the governing equations, initial and boundary conditions. This architecture can be used for solution and discovery (finding parameters) of systems of ordinary differential equations (ODEs) and partial differential equations (PDEs). While solving ODEs and PDEs with ANNs is not a new topic, e.g., \cite{Meade1994, Lagaris1998, Lagaris2000}, the success of these recent studies can be broadly attributed to: (1)~the choice of network architecture, i.e., the set of inputs and outputs of the ANN, so that one can impose governing equations on the network; (2)~algorithmic advances, including graph-based automatic differentiation for accurate differentiation of ANN functionals and for error back-propagation; and (3)~availability of advanced machine-learning software with CPU and GPU parallel processing capabilities including Theano \cite{bergstra2010theano} and TensorFlow \cite{abadi2016tensorflow}.

This framework has been used for solution and discovery of Schrodinger, Allen--Cahn, and Navier--Stokes equations \cite{Raissi2019, rudy2019data}. It has also been used for solution of high-dimensional stochastic PDEs \cite{han2018solving}. As pointed out in \cite{han2018solving}, this approach can be considered as a class of Reinforcement Learning \cite{Lange2012}, where the learning is on maximizing an incentive or minimizing a loss rather than direct training on data. If the network prediction does not satisfy a governing equation, it will result in an increase in the cost and therefore the learning traverses a path that minimizes that cost.

Here, we focus on the novel application of PINNs to solution and discovery of solid mechanics. We study linear elasticity in detail, but then illustrate the performance on nonlinear von~Mises elastoplasticity. Since parameters of the governing PDEs can also be defined as trainable parameters, the framework inherently allows us to perform parameter identification (model inversion). We validate the framework on synthetic data generated from low-order and high-order Finite Element Methods (FEM) and from Isogeometric Analysis (IGA) \cite{hughes2005isogeometric,cottrell2009isogeometric}. These datasets satisfy the governing equations with different order of accuracy, where the error can be considered as noise in data. We find that the training converges faster on more accurate datasets, pointing to importance of higher-order numerical methods for pre-training ANNs. We also find that if the data is pre-processed properly, the training converges to the correct solution and correct parameters even on data generated with a coarse mesh and low-order FEM---an important result that illustrates the robustness of the proposed approach. Finally, we find that, due to the imposition of the physics constraints, the training converges on a very sparse data set, which is a crucial property in practice given that the installation of a dense network of sensors can be very costly. 

Parameter estimation (identification) of complex models is a challenging task that requires a large number of forward simulations, depending on model complexity and the number of parameters. As a result, most inversion techniques have been applied to simplified models. The use of PINNs, however, allows us to perform identification simultaneously with fitting the ANN model on data \cite{Raissi2019}. This property highlights the potential of this approach compared with classical methods. We explore the application of PINN models for identification of multiple datasets generated with different parameters. Similar to transfer learning, where a pre-trained model is used as the initial state of the network \cite{taylor2009transfer}, we perform re-training on new datasets starting from a previously trained network on a different dataset (with different parameters). We find that the re-training and identification of other datasets take far less time. Since the successfully trained PINN model should also satisfy the physics constraints, it is in effect a surrogate model that can be used for extrapolation on unexplored data. To test this property, we train a network on four datasets with different parameters and then test it on a wide range of new parameter sets, and find that the results remain relatively accurate. This property points to the applicability of PINN models for sensitivity analysis, where classical approaches typically require an exceedingly large number of forward simulations.

\section{Physics-Informed Neural Networks: Linear Elasticity}\label{sec:elasticity}
In this section, we review the equations of linear elastostatics with emphasis on PINN implementation. 

\subsection{Linear elasticity}
The equations expressing momentum balance, the constitutive model and the kinematic relations are, respectively,
\begin{equation}\label{eqs:elas1}
\begin{split}
 \sigma_{ij,j} + f_i &=0, \\
 \sigma_{ij} &= \lambda \delta_{ij}\varepsilon_{kk} + 2\mu\varepsilon_{ij}, \\
 \varepsilon_{ij} &= \frac{1}{2}\left(u_{i,j} + u_{j,i}\right). 
\end{split}
\end{equation}
Here, $\sigma_{ij}$ denotes the Cauchy stress tensor. For the two-dimensional problems considered here $i,j=1,2$ (or $i,j=x,y$). We use the summation convention, and an subscript comma denotes partial derivative. The function $f_i$ denotes a body force, $u_i$ represents the displacements, $\varepsilon_{ij}$ is the infinitesimal stress tensor and $\delta_{ij}$ is the Kronecker delta. The Lam\'e parameters $\lambda$ and $\mu$ are the quantities to be inferred using PINN. 

\subsection{Introduction to Physics-Informed Neural Networks}\label{sec:pinn}
In this section, we provide an overview of the Physics-Informed Neural Networks (PINN) architecture, with emphasis on their application to model inversion. Let $\mathcal{N}(\mathbf{x}; \mathbf{W}, \mathbf{b}):\mathbb{R}^{d_{\mathbf{x}}}\rightarrow \mathbb{R}^{d_{\mathbf{y}}}$ be an $L$-layer neural network with input vector $\mathbf{x}$, output vector $\mathbf{y}$, and network parameters $\mathbf{W}, \mathbf{b}$. This network is a feed-forward network, meaning that each layer creates data for the next layer through the following nested transformations:
\begin{equation}\label{eqs:pinn1}
    \mathbf{z}^{l} = \sigma^{l} \left( \mathbf{W}^{l} \mathbf{z}^{l-1} + \mathbf{b}^{l} \right), \quad l=1,\dots,L,
\end{equation}
where $\mathbf{z}^0 \equiv \mathbf{x}$ and $\mathbf{z}^{L}\equiv\mathbf{y}$ are inputs and outputs of the model, $\mathbf{W}^l, \mathbf{b}^l$ are parameters of each layer~$l$, known as weights and biases, respectively. The functions $\sigma^l$ are called activation functions and make the network nonlinear with respect to the inputs. For instance, an ANN functional of some field variable, such as displacement $u(\mathbf{x})$, with three hidden layers and with $\sigma^l = \tanh$ as the activation function for all layers except the last can be written as 
\begin{equation}\label{eqs:pinn2}
\begin{split}
    \mathbf{z}^1(\mathbf{x}) &= \tanh(\mathbf{W}^0 \mathbf{x} + \mathbf{b}^0), \\
    \mathbf{z}^2(\mathbf{x}) &= \tanh(\mathbf{W}^1 \mathbf{z}^1 + \mathbf{b}^1), \\
    \mathbf{z}^3(\mathbf{x}) &= \tanh(\mathbf{W}^2 \mathbf{z}^2 + \mathbf{b}^2),  \\
    u(\mathbf{x}) &= \mathbf{W}^3 \mathbf{z}^3 + \mathbf{b}^3.
\end{split}
\end{equation}
This model can be considered as an approximate solution for the field variable~$u$ of a partial differential equation.

In the PINN architecture, the network inputs (also known as features) are space and time variables, i.e., $(x,y,z,t)$ in Cartesian coordinates, which makes it meaningful to perform the differentiation of the network's output with respect to any of the input variables. Classical implementations based on finite difference approximations are not accurate when applied to deep networks (see \cite{baydin2017automatic} for a review). Thanks to modern graph-based implementation of the feed-forward network (e.g., Theano \cite{bergstra2010theano}, Tensorflow \cite{abadi2016tensorflow}, MXNet \cite{chen2015mxnet}), this can be carried out using Automatic Differentiation at machine precision, therefore allowing for many hidden layers to represent nonlinear response. Hence, evaluation of a partial differential operator $\mathcal{P}$ acting on $u$ is achieved naturally with graph-based differentiation and can then be incorporated in the cost function along with initial and boundary conditions as:
\begin{equation}\label{eqs:pinn3}
    \mathcal{L}=|u - u^*| + |\mathcal{P}u - 0^*| + |u-u^*|_{\partial \Omega} + |u_0-u_0^*|,
\end{equation}
where $\partial \Omega$ is the domain boundary, $u_0-u_0^*$ is the initial condition at $t=t_0$, and $0^*$ indicates the expected (true) value for the differential relation $\mathcal{P}u$ at any given training point. The norm $|\cdot|$ of a generic quantity $g$ defined in $\Omega$ denotes $\frac{1}{N}\sum_{i=1}^{N}g(\mathbf{x}_i)^2$ where the $\mathbf{x}_i$'s are the spatial points where the data is known. The dataset is then fed to the neural network and an optimization is performed to evaluate all the parameters of the model, including the parameters of the PDE.

\subsection{Training PINN}
Different algorithms that can be used to train a neural network. Among the choices available in Keras \cite{chollet2015keras} we use the Adam optimization scheme \cite{kingma2014adam}, which we have found to outperform other choices such as Adagrad \cite{duchi2011adaptive}, for this task. Several algorithmic parameters affect the rate of convergence of the network training. Here we adopt the terminology in Keras \cite{chollet2015keras}, but the terminology in other modern machine learning packages is similar. The algorithmic parameters include \emph{batch-size}, \emph{epochs}, \emph{shuffle}, and \emph{patience}. Batch-size controls the number of samples from a dataset used to evaluate one gradient update. A batch-size of~1 would be associated with a full stochastic gradient descent optimization. One epoch is one round of training on a dataset. If a dataset is shuffled, then a new round of training (epoch) would result in an updated parameter set because the batched-gradients are evaluated on different batches. It is common to re-shuffle a dataset many times and perform the back-propagation updates. The optimizer may, however, stop earlier if it finds that new rounds of epochs are not improving the cost function. That is where the last keyword, patience, comes in. This is mainly because we are dealing with non-convex optimization and we need to test the training from different starting points and in different directions to build confidence on the parameters evaluated from minimization of the cost-function on a dataset. Patience is the parameter that controls when the optimizer should stop the training. 

There are three ways to train the network: (1)~generate a sufficiently large number of datasets and perform a one-epoch training on each dataset, (2)~work on one dataset over many epochs by reshuffling the data, and (3)~a combination of these. When dealing with synthetic data, all approaches are feasible to pursue. However, strategy~(1) above is usually impossible to apply in practice, specially in space, where sensors are installed at fixed and limited locations. In the original work on PINN \cite{Raissi2019}, approach~(1) was used to train the model, where datasets are generated on random space discretizations at each epoch. Here, we follow approach~(2) to use training data that we could realistically have in practice. For all examples, unless otherwise noted, we use a batch-size of 64, a limit of 10,000 epochs with shuffling, and a patience of 500 to perform the training.

\section{Illustrative Example and Discussion}
In this section, we use the PINN architecture on an illustrative linear elasticity problem. 

\subsection{Problem setup}
\begin{figure}[!ht]
	\centering
	\includegraphics[trim={0.0in 0.2in 0.0in 0.2in}, clip, width=0.5\textwidth]{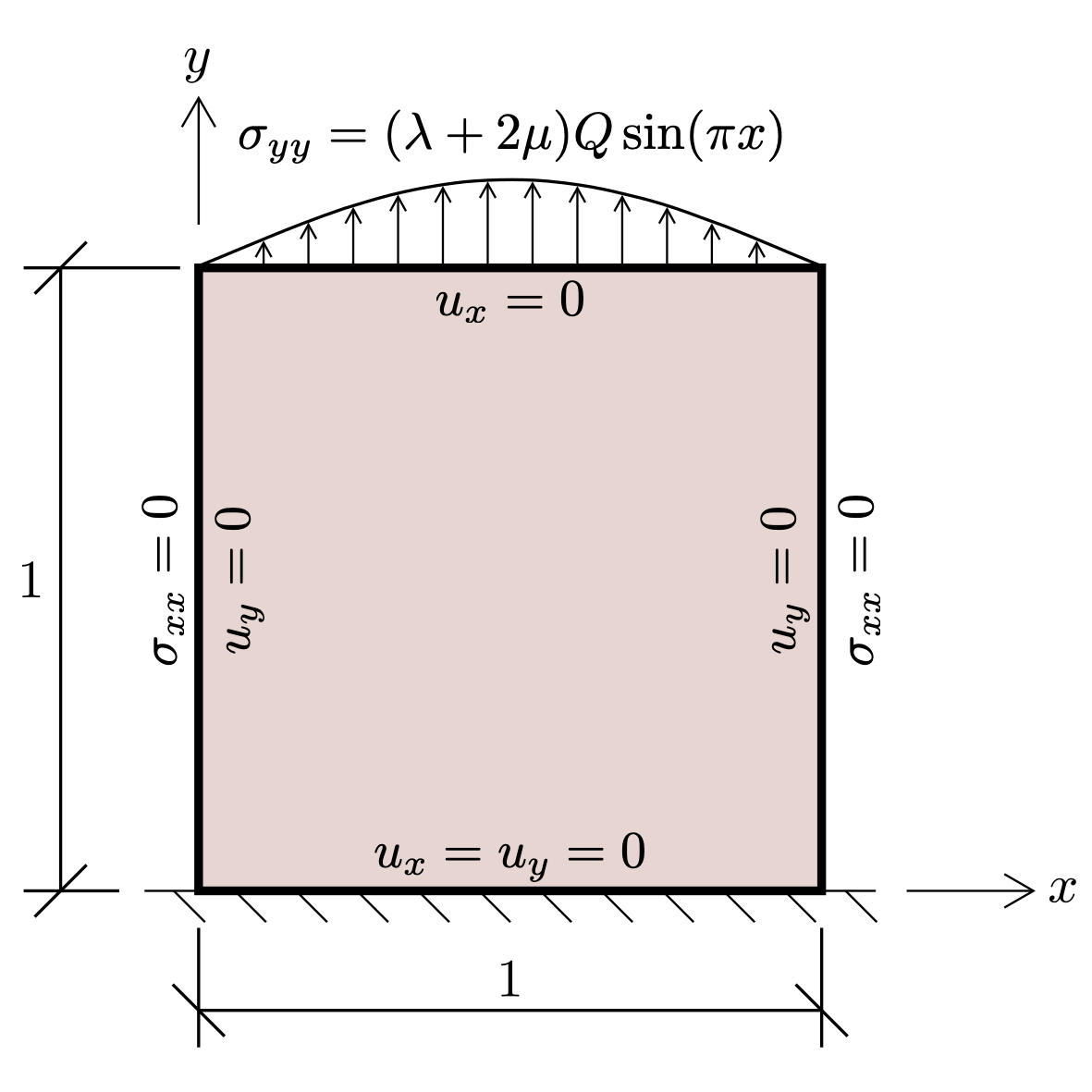}
	\caption{Problem setup and boundary conditions.}
	\label{fig:problem_setup}
\end{figure}

\begin{figure}[!ht]
    \centering
    \includegraphics[width=1\textwidth]{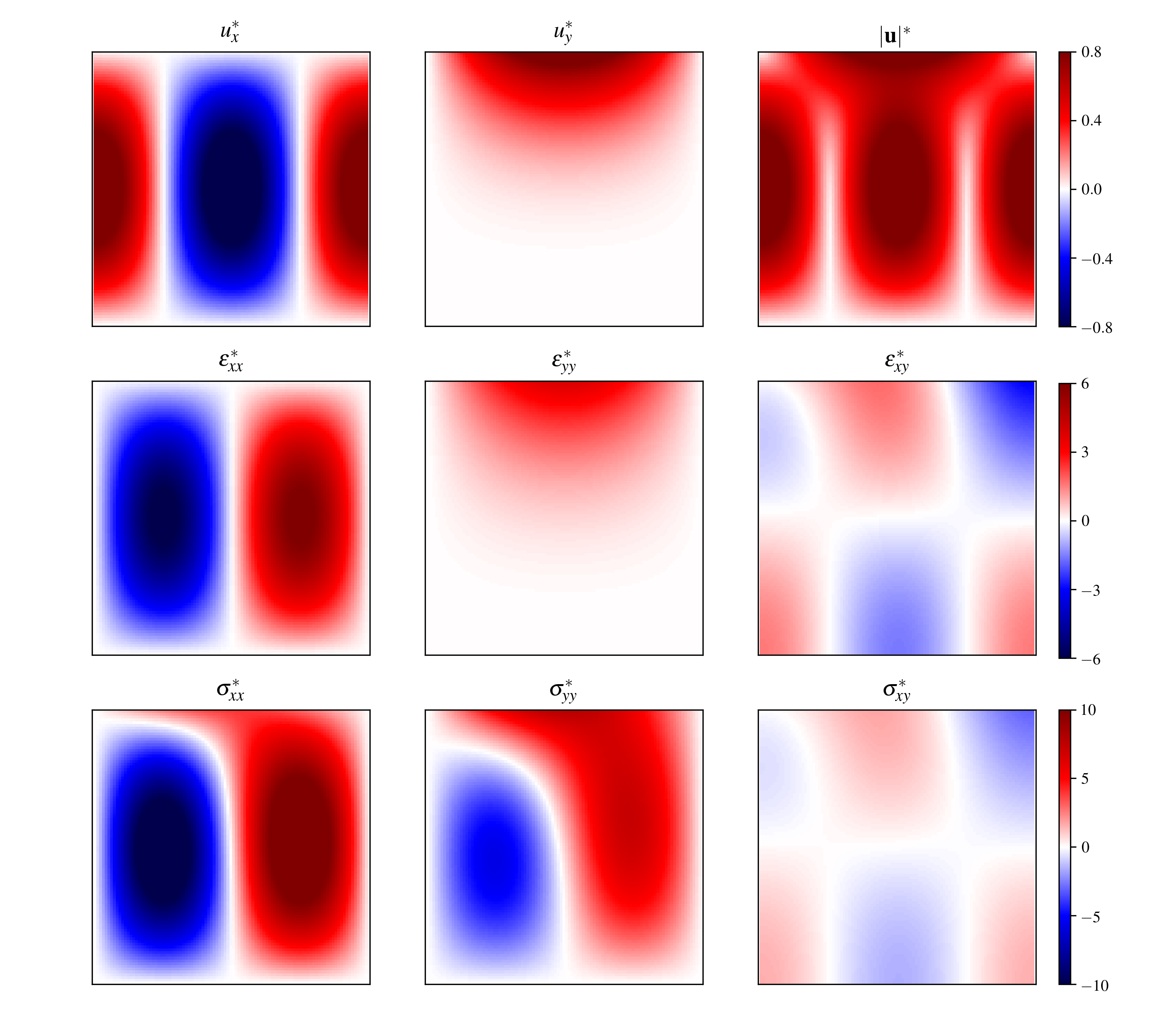}
    \caption{Exact solution in Eqs.~\eqref{eqs:results2a}--\eqref{eqs:results2b} for parameter values of $\lambda=1$, $\mu=0.5$, and $Q=4$.}
    \label{fig:analytical_sol}
\end{figure}

To illustrate the application of the proposed approach, we consider an elastic plane-strain problem on the unit square (Fig.~\ref{fig:problem_setup}), subject to the boundary conditions depicted in the figure. The body forces are:
\begin{equation}\label{eqs:results1}
\begin{split}
f_x &= \lambda \left[4\pi^2 \cos(2\pi x) \sin(\pi y) - \pi \cos(\pi x) Q y^3 \right]  				 \\
  ~ &+ \mu \left[9\pi^2 \cos(2\pi x) \sin(\pi y) - \pi \cos(\pi x) Q y^3 \right]       				 \\
f_y &= \lambda \left[-3\sin(\pi x) Qy^2 +2\pi^2 \sin(2\pi x) \cos(\pi y) \right]         			 \\
  ~ &+ \mu \left[-6\sin(\pi x) Qy^2 +2\pi^2\sin(2\pi x)\cos(\pi y) + \pi^2\sin(\pi x) Q y^4/4 \right]. 
\end{split}
\end{equation}
The exact solution of this problem is 
\begin{alignat}{2}\label{eqs:results2a}
u_x(x,y)&=\cos(2\pi x) \sin(\pi y), \\
u_y(x,y)&=\sin(\pi x) Qy^4/4. \label{eqs:results2b}
\end{alignat}
which is plotted in Fig.~\ref{fig:analytical_sol}, for parameter values of $\lambda=1$, $\mu=0.5$, and $Q=4$. 

\subsection{Neural Network setup}\label{sec:neural_network}
\begin{figure}[!ht]
	\centering
	\includegraphics[trim={0.0in 0.0in 0.0in 0.0in}, clip, width=1.0\textwidth]{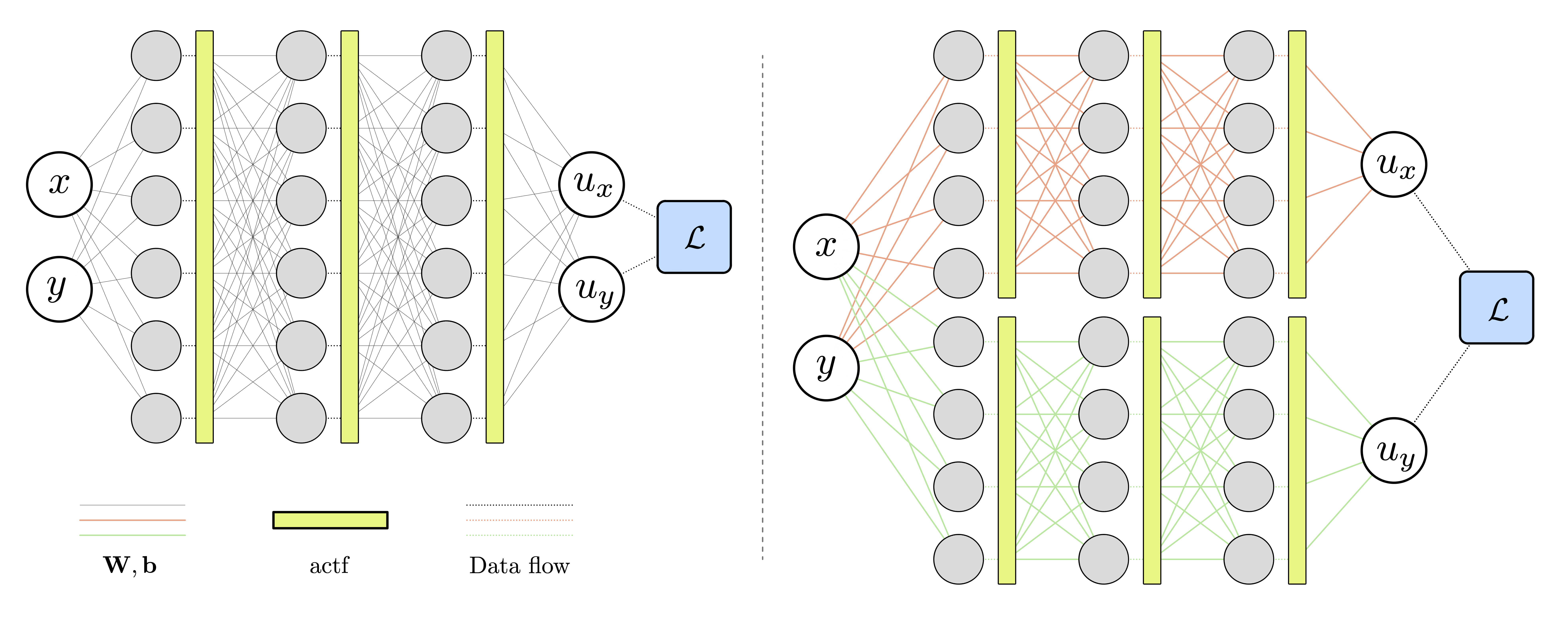}
	\caption{Potential PINN network choices, with $u_x$ and $u_y$ as outputs of a single network (left), or outputs of two independent networks with different parameters (right).}
	\label{fig:net_choices}
\end{figure}

\begin{figure}[!ht]
    \centering
    \includegraphics[width=0.7\textwidth]{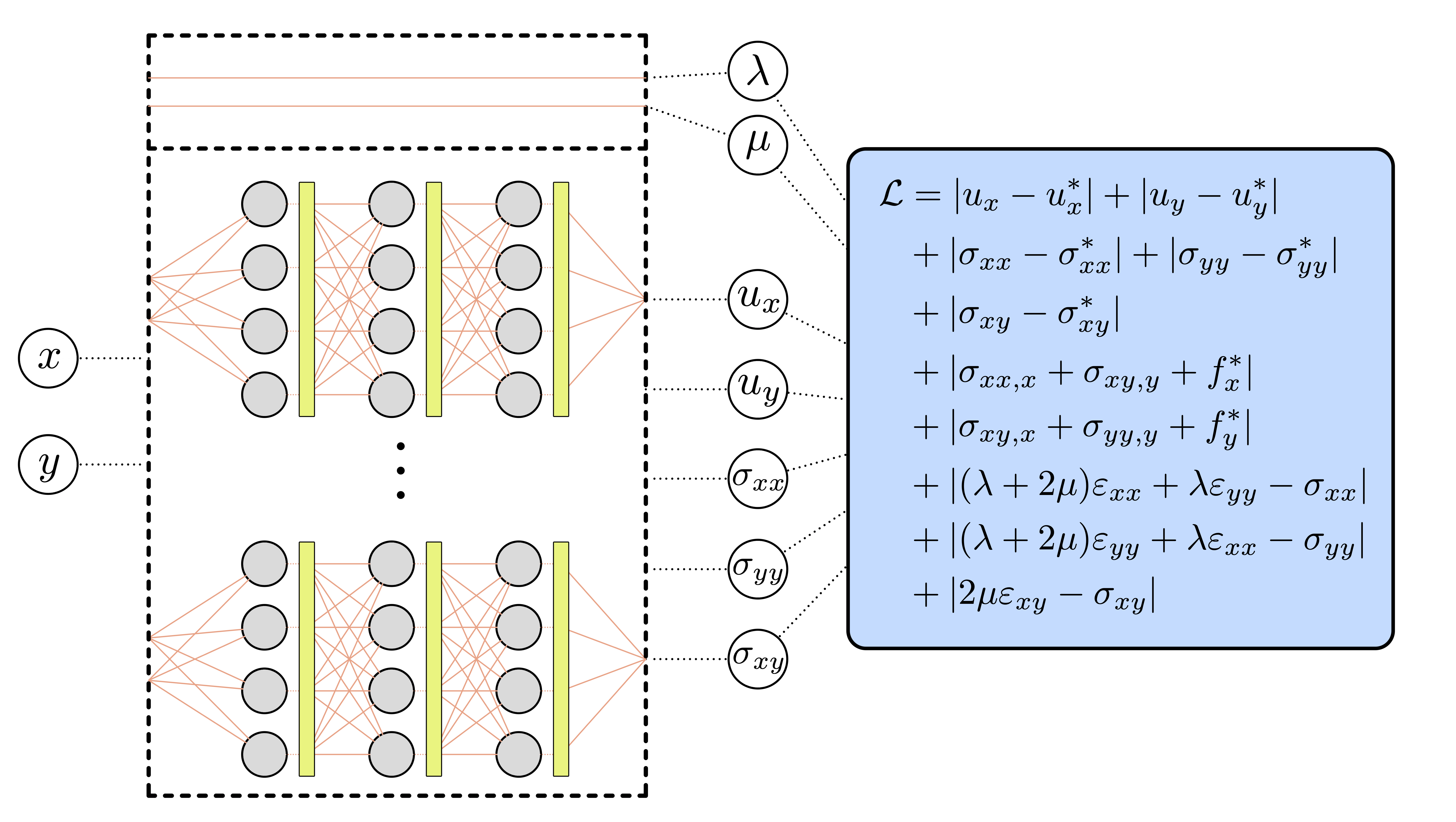}
    \caption{Network architecture of choice used in this study. We define five networks, one for each variable of interest, i.e., $u_x, u_y, \sigma_{xx}, \sigma_{yy}, \sigma_{xy}$. Each network has $(x,y)$ as input features. }
    \label{fig:neural_network_arch}
\end{figure}

Due to the symmetry of the stress and strain tensors, the quantities of interest for a two-dimensional problem are $u_x$, $u_y$, $\varepsilon_{xx}$, $\varepsilon_{yy}$, $\varepsilon_{xy}$, $\sigma_{xx}$, $\sigma_{yy}$, $\sigma_{xy}$. There are a few potential architectures that we can use to design our network. The input features (variables) are the spatial coordinates $(x,y)$, for all the network choices. For the outputs, a potential design is to have a densely connected network with two outputs as $(u_x,u_y)$. Another option is to have two densely connected independent networks with only one output each, associated with $u_x$ and $u_y$, respectively (Fig. \ref{fig:net_choices}). Then, the remaining quantities of interest, i.e., $\sigma_{ij},\varepsilon_{ij}$, can be obtained through differentiation. Alternatively, we may have $(u_x,u_y,\sigma_{xx},\sigma_{yy},\sigma_{xy})$ or $(u_x,u_y,\varepsilon_{xx},\varepsilon_{yy},\varepsilon_{xy})$ as outputs of one network or multiple independent networks. As can be seen from Fig.~\ref{fig:net_choices}, these choices affect the number of parameters of the network and how different quantities of interest are correlated. Equation~\eqref{eqs:pinn2} shows that the the feed-forward neural network imposes a special functional form to the network that may not necessarily follow any cross-dependence between variables in the governing equations \eqref{eqs:elas1}. Our data shows that using separate networks for each variable results in a far more effective strategy. Therefore, we propose to have variables $u_x,u_y,\sigma_{xx},\sigma_{yy},\sigma_{xy}$ defined as independent ANNs as our architecture of choice (see Fig. \ref{fig:neural_network_arch}), i.e.
\begin{equation}\label{eqs:nn0}
\begin{split}
&u_x(\mathbf{x}) \approx \mathcal{N}_{u_x}(\mathbf{x}), \\
&u_y(\mathbf{x})\approx\mathcal{N}_{u_y}(\mathbf{x}),   \\
&\sigma_{xx}(\mathbf{x}) \approx \mathcal{N}_{\sigma_{xx}}(\mathbf{x}), \\
&\sigma_{yy}(\mathbf{x}) \approx \mathcal{N}_{\sigma_{yy}}(\mathbf{x}), \\
&\sigma_{xy}(\mathbf{x}) \approx \mathcal{N}_{\sigma_{xy}}(\mathbf{x}),
\end{split}
\end{equation}

The cost function is defined as
\begin{equation}\label{eqs:nn}
\begin{split}
\mathcal{L} &= |u_x-u_x^*| + |u_y-u^*_y| + |\sigma_{xx}-\sigma_{xx}^*| + |\sigma_{yy}-\sigma_{yy}^*| + |\sigma_{xy}-\sigma_{xy}^*| \\
            &+|\sigma_{xx,x} + \sigma_{xy,y} + f_x^*| + |\sigma_{xy,x} + \sigma_{yy,y} + f_y^*|        \\
            &+ |(\lambda+2\mu)\varepsilon_{xx} + \lambda \varepsilon_{yy} - \sigma_{xx}| + |(\lambda+2\mu) \varepsilon_{yy} + \lambda \varepsilon_{xx} - \sigma_{yy}| + |2\mu\varepsilon_{xy} - \sigma_{xy}|.    
\end{split}
\end{equation}
The quantities with asterisks represent given data. We will train the networks so that their output values are as close as possible to the data, which may be real field data or, in this paper, synthetic data from the exact solution to the problem or the result of a high-fidelity simulation. The values without asterisk represent either direct outputs of the network (e.g., $u_x$ or $\sigma_{xx}$; see Eq.~\eqref{eqs:nn0}) or quantities obtained through automatic graph-based differentiation \cite{baydin2017automatic} of the network outputs (e.g., $\varepsilon_{xx}=u_{x,x}$). In Eq.~\eqref{eqs:nn}, $f_x^*$ and $f_y^*$ represent data on the body forces obtained as $f_i^*=-\sigma_{ij,j}^*$.

The different terms in the cost function represent measures of the error in the displacement and stress fields, the momentum balance, and the constitutive law. This cost function can be used for deep-learning-based solution of PDEs as well as for identification of the model parameters. For the solution of PDEs, $\lambda$ and $\mu$ are treated as fixed numbers in the network. For parameter identification, $\lambda$ and $\mu$ are treated as network parameters that change during the training phase (see Fig.~\ref{fig:neural_network_arch}). In TensorFlow \cite{abadi2016tensorflow} this can be accomplished defining $\lambda$ and $\mu$ as \texttt{Constant} (PDE solution) or \texttt{Variable} (parameter identification) objects, respectively. We set up the problem using the SciANN \cite{haghighat2019sciann} framework, a high-level Keras \cite{chollet2015keras} wrapper for physics-informed deep learning and scientific computations. Experimenting with all of the previously mentioned network choices can be easily done in SciANN with minimal coding.\footnote{The code for some of the examples solved here is available at: \href{https://github.com/sciann/examples}{https://github.com/sciann/examples}.}

\subsection{Identification of model parameters: PINN trained on the exact solution}\label{sec:pinn_analytical}

Here, we use PINN to identify the model parameters $\lambda$ and $\mu$. Our data corresponds to the exact solution with parameter values $\lambda=1$, $\mu=0.5$ and $Q=4$. Our default dataset consists of 100$\times$100 sample points, uniformly distributed. We study how the accuracy and the efficiency of the identification process depend on the architecture and functional form of the network; the available data; and whether we use one or several independent networks for the different quantities of interest. To study the impact of the architecture and functional form of the ANN, we use 4 different networks with either 5 or 10 hidden layers, and either 20 or 50 neurons per layer; see Table~\ref{table1}. The role of the network functional form is studied comparing the performance of the two most widely used activation functions, i.e., $\tanh$ and ReLU, where $\textrm{ReLU}(x)=\max(0,x)$ \cite{bishop2006pattern}.

Studying the impact of the available data on the identification process is crucial because we are interested in identifying the model parameters with as little data as possible. We undertake the analysis considering two scenarios:
\begin{enumerate}
\item[(a)] \emph{Stress-complete data}: In this case, we have data at a set of points for the displacements and their first-order derivatives, that is, $u_x^*$, $u_y^*$, $\sigma_{xx}^*$, $\sigma_{yy}^*$, $\sigma_{xy}^*$. Because our cost function \eqref{eqs:nn} involves also data that depends on the stress derivatives ($f_x^*$ and $f_y^*$), this approach relies on an additional algorithmic procedure for differentiation of stresses. In this section we compute the stress derivatives using second-order central finite-difference approximations.
\item[(b)] \emph{Force-complete data}: In this scenario, we have data at a set of points for the displacements, their first derivatives and their second derivatives. The availability of the displacement second derivatives allows us to determine data for the body forces $f_x^*$ and $f_y^*$ using the momentum balance equation without resorting to any differentiation algorithm.
\end{enumerate}

\begin{table}
	\centering
	{\small 
	\caption{Statistics of the networks of choice to perform PINN learning.}\label{table1}
	\begin{tabular}{ccccc}
		\hline
		\multirow{2}{*}{Network} & \multirow{2}{*}{Layers} & \multirow{2}{*}{Neurons} & \multicolumn{2}{c}{Number of Parameters}        \\
		& & & Independent Networks & Single Network \\
		\hline
		i    & 5                       & 20                       & 12336                & 1893           \\
		ii   & 5                       & 50                       & 72816                & 10713          \\
		iii  & 10                      & 20                       & 27036                & 3993           \\
		iv   & 10                      & 50                       & 162066               & 23463          \\
		\hline
	\end{tabular}
}
\end{table}

\begin{figure}[!ht]
	\centering
	\includegraphics[width=0.8\textwidth]{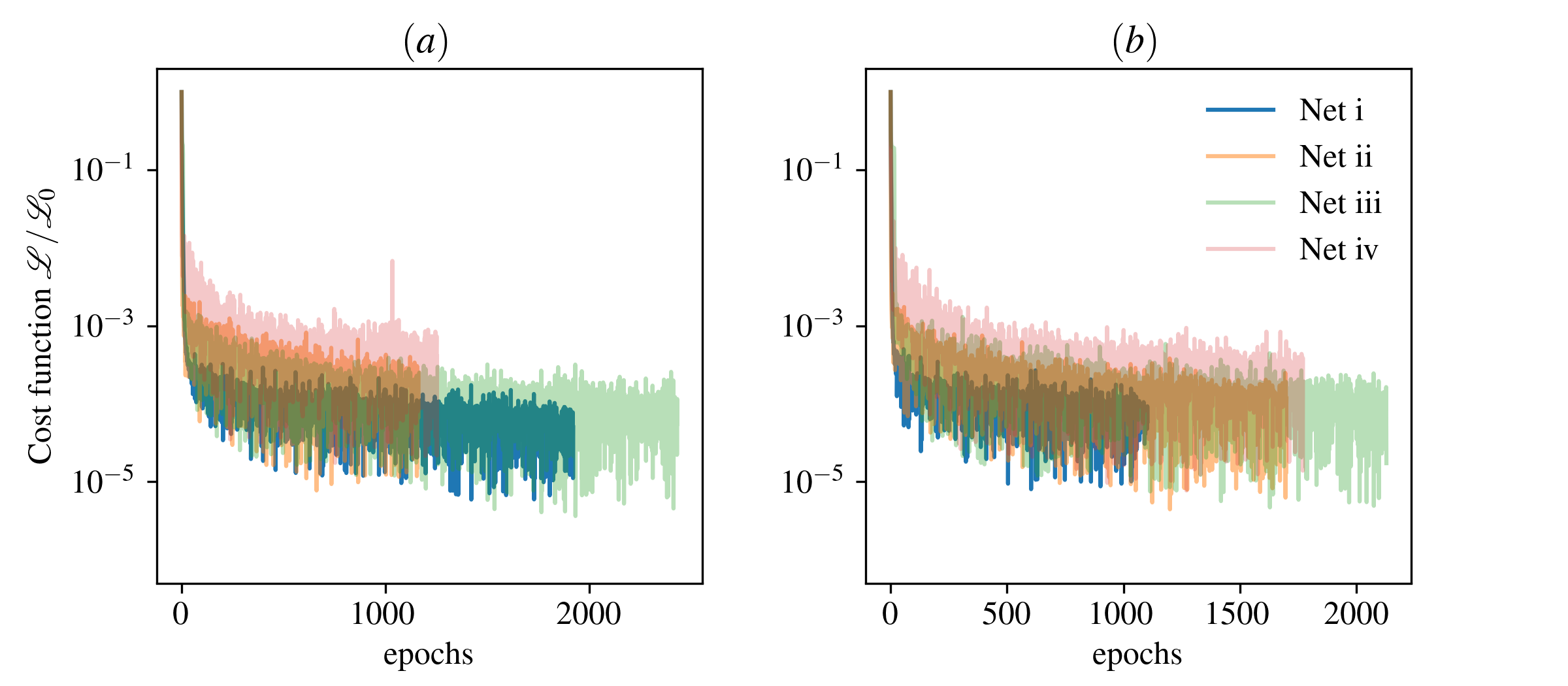}
	\caption{The result of training networks i, ii, iii, and iv on the analytical data set $u_x$, $u_y$, $\sigma_{xx}$, $\sigma_{yy}$, and $\sigma_{xy}$; (a)~body forces are evaluated from central-difference differentiation of stress components, (b)~body forces are also given analytically.}
	\label{fig:pinn_analytical_ab}
\end{figure}

In Fig.~\ref{fig:pinn_analytical_ab} we compare the evolution of the cost function for stress-complete data (Fig.~\ref{fig:pinn_analytical_ab}a) and force-complete data (Fig.~\ref{fig:pinn_analytical_ab}b). Both figures show a comparison of the four network architectures that we study; see Table~\ref{table1}. We find that training on the force-complete data performs slightly better (lower loss) at a given epoch.

\begin{figure}[!ht]
	\centering
	\includegraphics[width=0.8\textwidth]{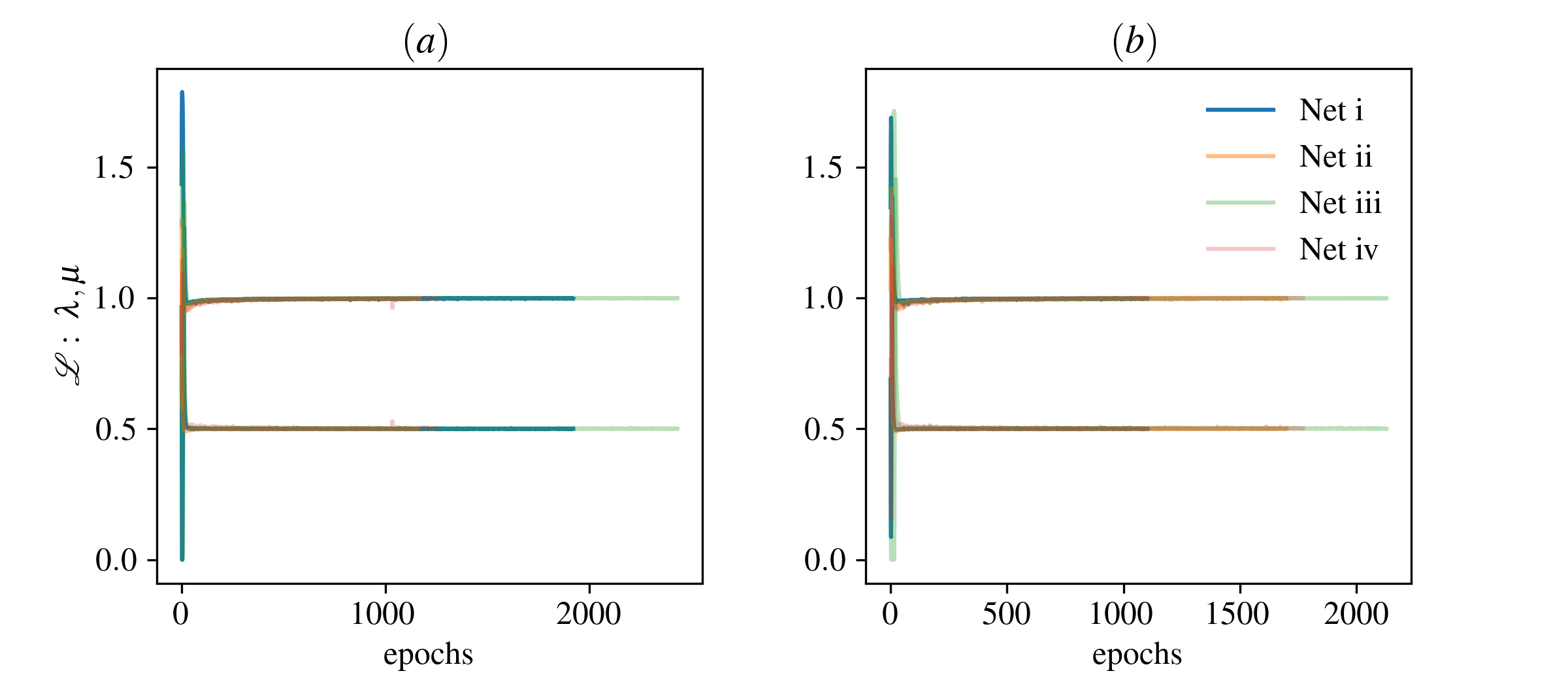}
    \includegraphics[width=0.8\textwidth]{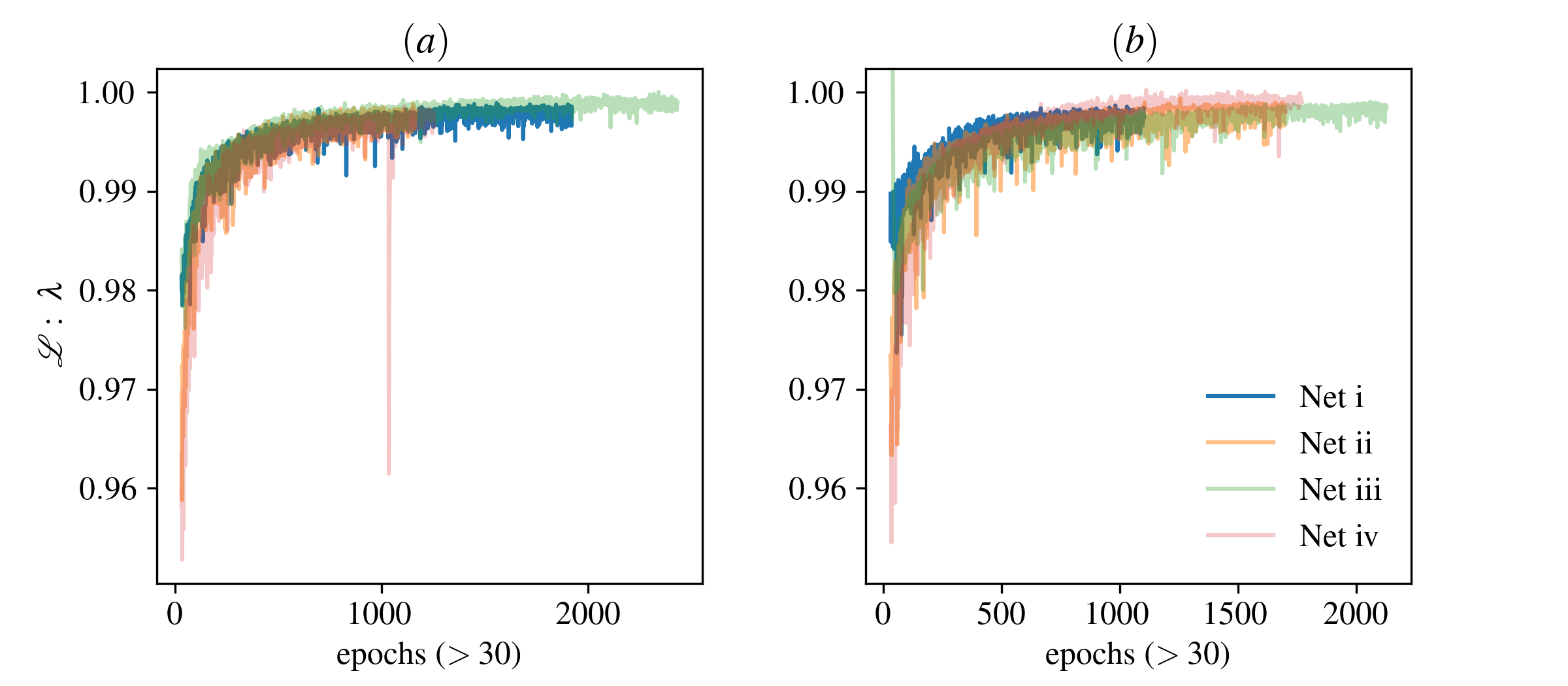}
    \includegraphics[width=0.8\textwidth]{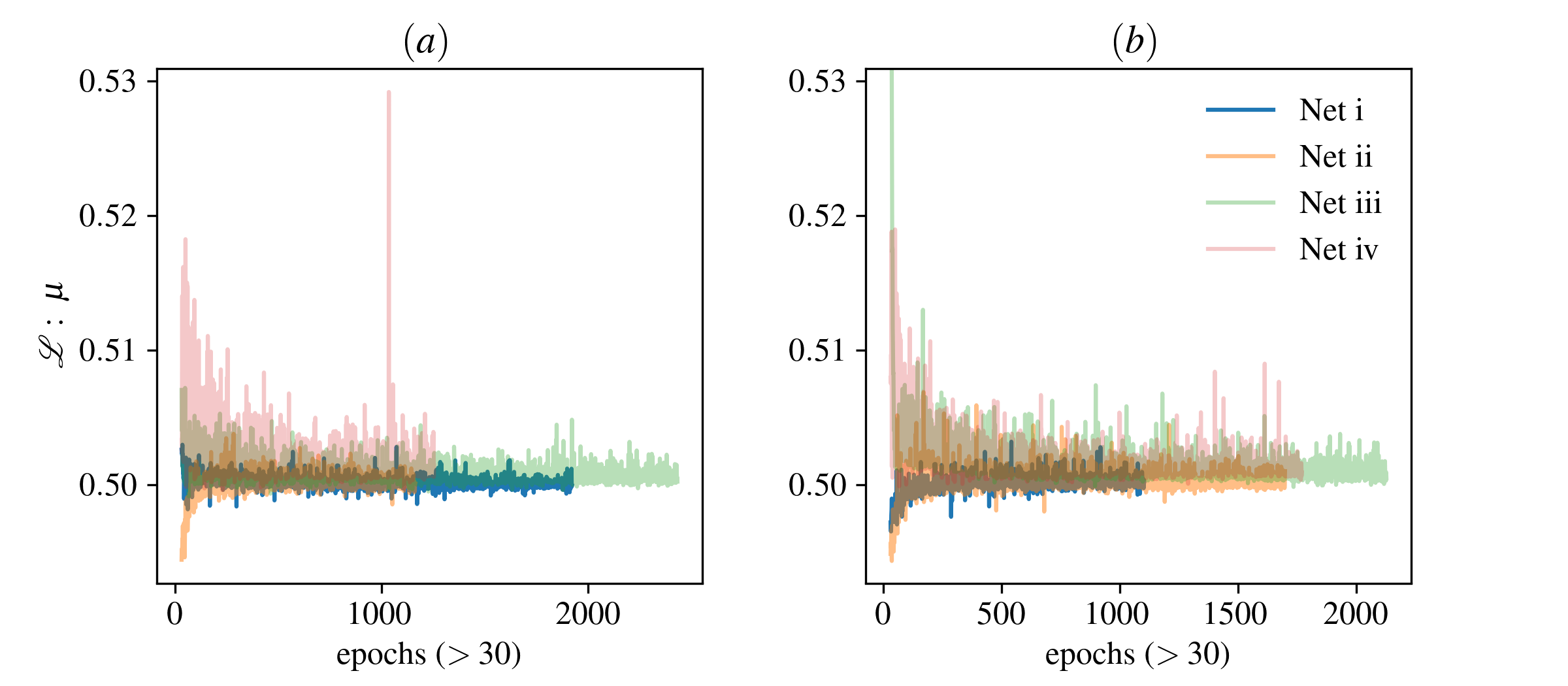}
    \caption{The result of identification for $\lambda=1, \mu=1/2$ for networks i, ii, iii, and iv on the analytical data set $u_x$, $u_y$, $\sigma_{xx}$, $\sigma_{yy}$, and $\sigma_{xy}$; (a)~body forces are evaluated from central-difference differentiation of stress components, (b)~body forces are also given analytically.}
	\label{fig:pinn_analytical_ab_lames}
\end{figure}

\begin{figure}[!ht]
	\centering
	\includegraphics[width=0.8\textwidth]{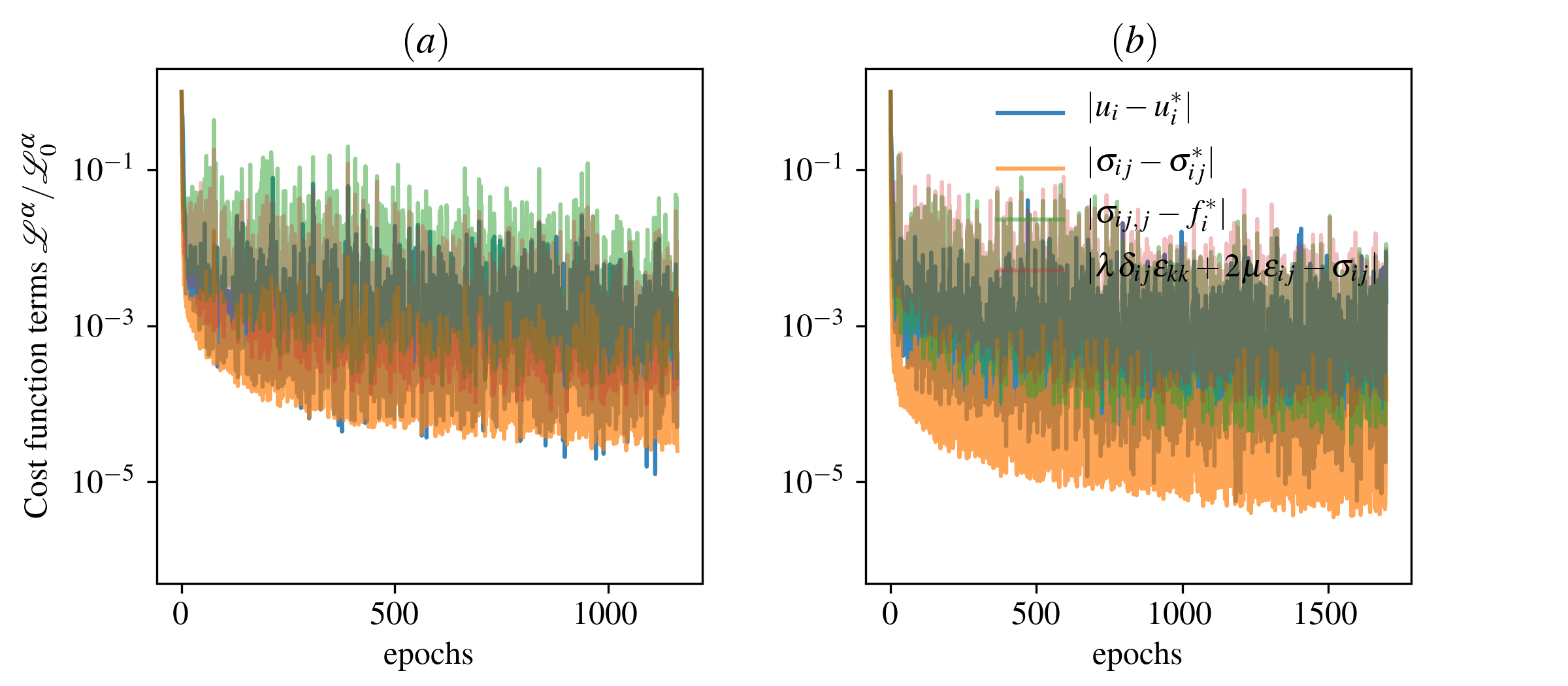}
	\caption{Individual terms of total loss \eqref{eqs:nn} for network~ii on the analytical data set $u_x$, $u_y$, $\sigma_{xx}$, $\sigma_{yy}$, and $\sigma_{xy}$; (a)~body forces are evaluated from central-difference differentiation of stress components, (b)~body forces are also given analytically.}
	\label{fig:pinn_analytical_ab_all_terms}
\end{figure}

The result of convergence of model identification is shown in Fig.~\ref{fig:pinn_analytical_ab_lames}. The training converges to the true values of parameters, i.e., $\lambda=1$ and $\mu=1/2$, for all cases. We find that the optimization is very quick on the parameters while it takes far more epochs to fit the network on the field variables. Additionally, we observe that deeper networks produce less accurate parameters. We attribute the loss of accuracy as we increase the ANN complexity to over-fitting \cite{bishop2006pattern,Goodfellow2016}. Convergence of the individual terms in the loss function \eqref{eqs:nn} is shown in Fig.~\ref{fig:pinn_analytical_ab_all_terms} for Net-ii (see Table \ref{table1}). We find that all terms in the loss, i.e., data-driven and physics-informed, show oscillations during the optimization. Therefore, no individual term is solely responsible for the oscillations in the total loss (Fig.~\ref{fig:pinn_analytical_ab}).

\begin{figure}[!ht]
	\centering
	\includegraphics[width=0.8\textwidth]{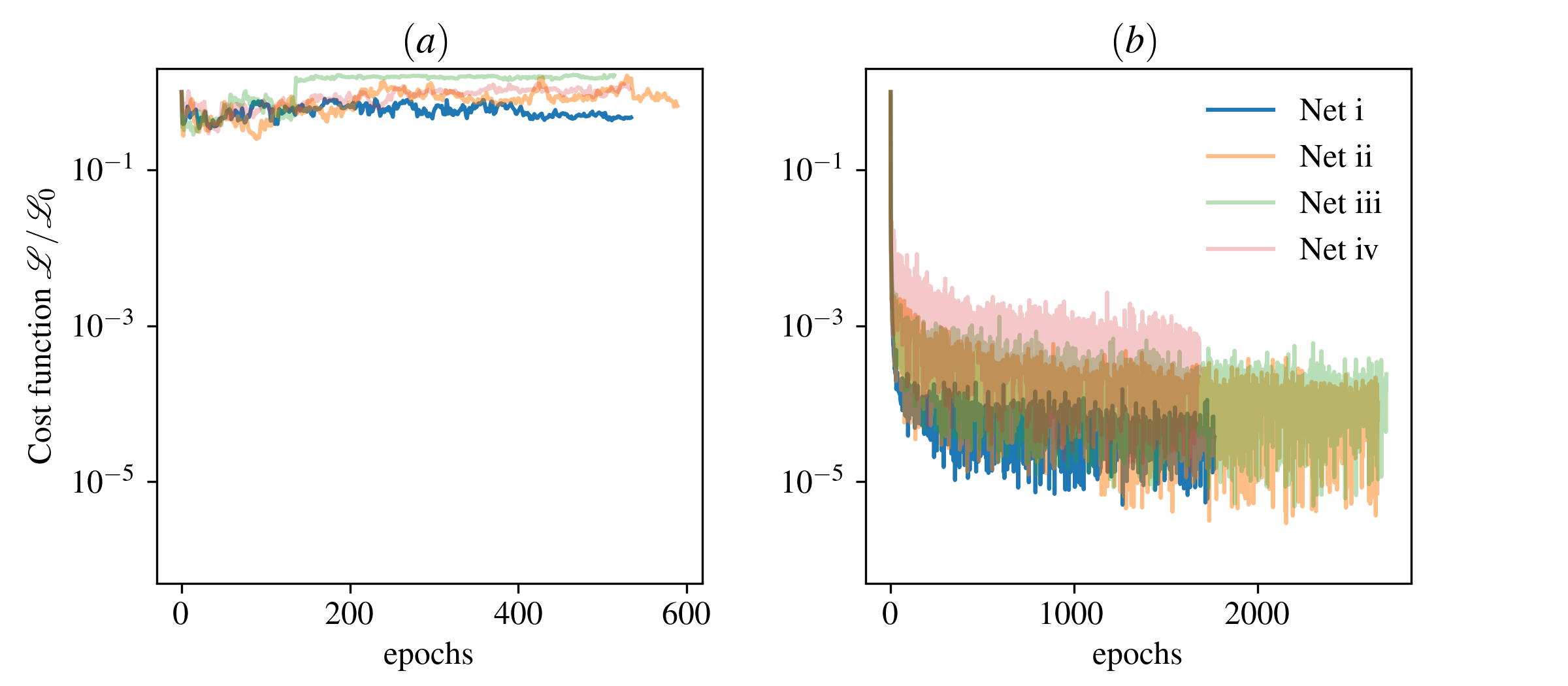}
	\caption{(a)~$\textrm{ReLU}$ activation function on the analytical data set $u_x$, $u_y$, $\sigma_{xx}$, $\sigma_{yy}$, $\sigma_{xy}$, $f_x$ , and $f_y$. (b)~Connected network. }
	\label{fig:pinn_analytical_relu_tanh}
\end{figure}

The impact of the ANN functional form can be examined comparing the data in Figs.~\ref{fig:pinn_analytical_ab}b and \ref{fig:pinn_analytical_relu_tanh}a, which show the evolution of the cost function using the activation functions $\tanh$ and $\textrm{ReLU}$, respectively. The function $\textrm{ReLU}$ has discontinuous derivatives, which explains its poor performance for physics-informed deep learning, whose effectiveness relies heavily on accurate evaluation of derivatives.

A comparison of Figs.~\ref{fig:pinn_analytical_ab}b and \ref{fig:pinn_analytical_relu_tanh}b shows that using independent networks for displacements and stresses is more effective than using a single network. We find that the single network leads to less accurate elastic parameters because the cross-dependencies of the network outputs through the kinematic and constitutive relations may not be adequately represented by the $\tanh$ activation function.

\begin{figure}[!ht]
    \centering
    \includegraphics[width=0.8\textwidth]{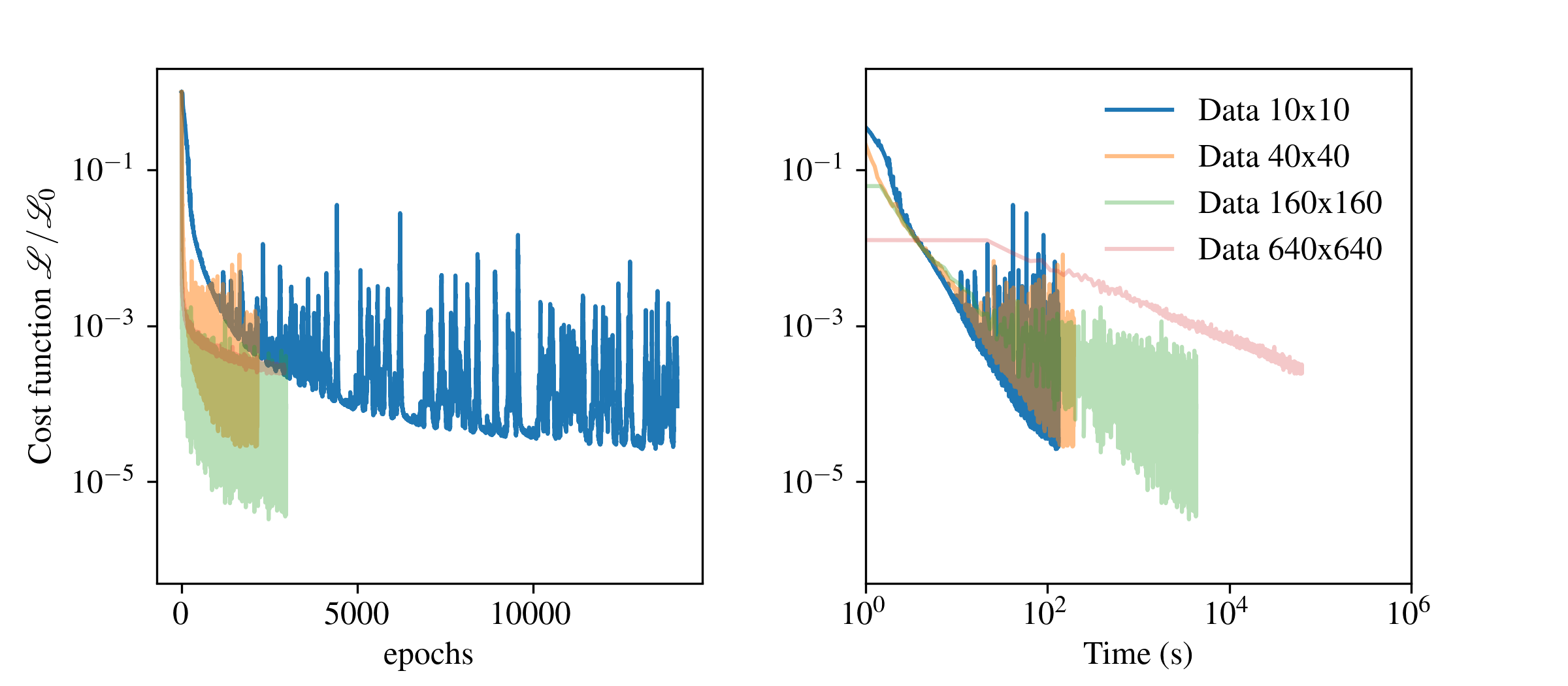}
    \caption{Training on different sizes of data. Parameters are all accurately identified however solution has different level of accuracy. }
    \label{fig:pinn_analytical_griddata}
\end{figure}

Fig.~\ref{fig:pinn_analytical_griddata} analyzes the effect of availability of data on the training. We computed the exact solution on four different uniform grids of size $10\times10$, $40\times40$, $160\times160$, and $640\times640$; and carried out the parameter identification process. We performed the comparison using force-complete data and a network with 10~layers and 20~neurons per layer (network~iii). The training process found good approximations to the parameters for all cases, including that with only $10\times10$~points.  The results show that fewer data points require many more epoch cycles, but the overall computational cost is far lower. 

\subsection{PINN models trained on the FEM solution}\label{sec:pinn_fem}

Here, we generate synthetic data from FEM solutions, and then perform the training. The domain is discretized with a mesh comprised of $40\times 40$ elements. Four datasets are prepared using quadrilateral bilinear, biquadratic, bicubic, and biquartic Lagrange $C^0$ elements using the commercial FEM software COMSOL~\cite{COMSOL}. We evaluate the FEM displacements, strains, stresses and stress derivatives at the center of each element. Then, we map the data to a $100\times100$ training grid using SciPy's griddata module with cubic interpolation. This step is performed as a data-augmentation procedure, which is a common practice in machine learning \cite{bishop2006pattern}. 

\begin{figure}[!ht]
    \centering
    \includegraphics[width=0.8\textwidth]{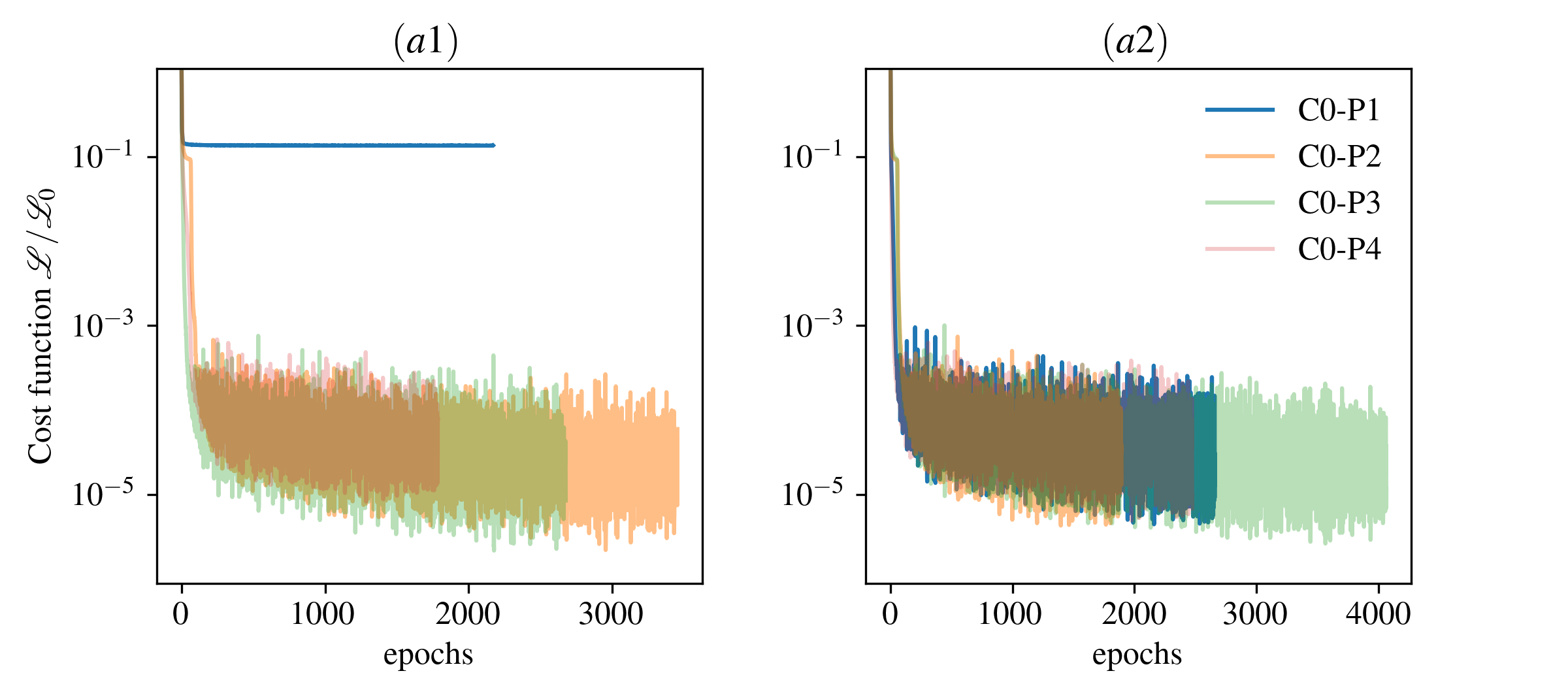}
    \caption{(a1)~Training on the FEM dataset using $u_x$, $u_y$, $\sigma_{xx}$, $\sigma_{yy}$, $\sigma_{xy}$, $f_x$ and $f_y$ components. (a2)~Training with body forces $f_x$ and $f_y$ evaluated from central-differentiation of stress components.}
    \label{fig:pinn_fem_a1}
\end{figure}

\begin{figure}[!ht]
    \centering
    \includegraphics[width=1\textwidth]{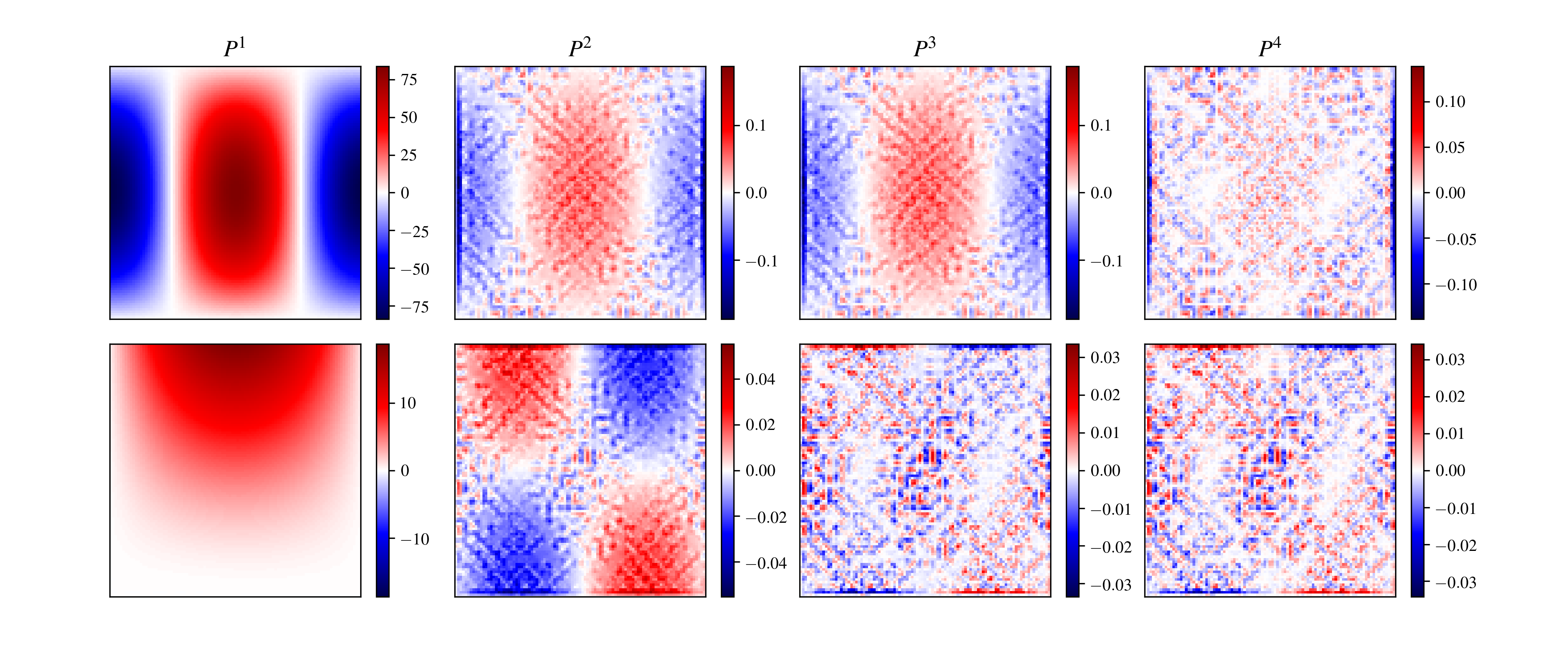}
    \caption{The error in bilinear, biquadratic, bicubic, and biquartic FEM data, that is evaluated as the difference between FEM evaluation of momentum relation, i.e., $\sigma_{ij,j}$ and true body forces $f_i^*$ in $x$ (top) and $y$ (bottom) directions.}
    \label{fig:pinn_fem_a1_error}
\end{figure}

To analyze the importance of data satisfying the governing equations of the system, we focus our attention on network~ii and we study cases with stress- and force-complete data. The results of training are presented in Fig.~\ref{fig:pinn_fem_a1}. As can be seen here, the bilinear element performs poorly on the learning and identification. The performance of training on the other elements is good, comparable to that using the analytical solution. Further analysis shows that this is indeed expected as FEM differentiation of bilinear elements provides a poor approximation of the body forces. The error in the body forces is shown in Fig.~\ref{fig:pinn_fem_a1_error}, which indicates a high error for bilinear elements. We conclude that the standard bilinear elements are not suitable for this problem to generate numerical data for deep learning. Fig.~\ref{fig:pinn_fem_a1}a2 confirms that pre-processing the data can remove the error that was present in the numerical solution with bilinear elements, and enable the optimization to successfully complete the identification.

\subsection{PINN models trained on the IGA solution}\label{sec:pinn_iga}

\begin{figure}[!ht]
    \centering
    \includegraphics[width=0.8\textwidth]{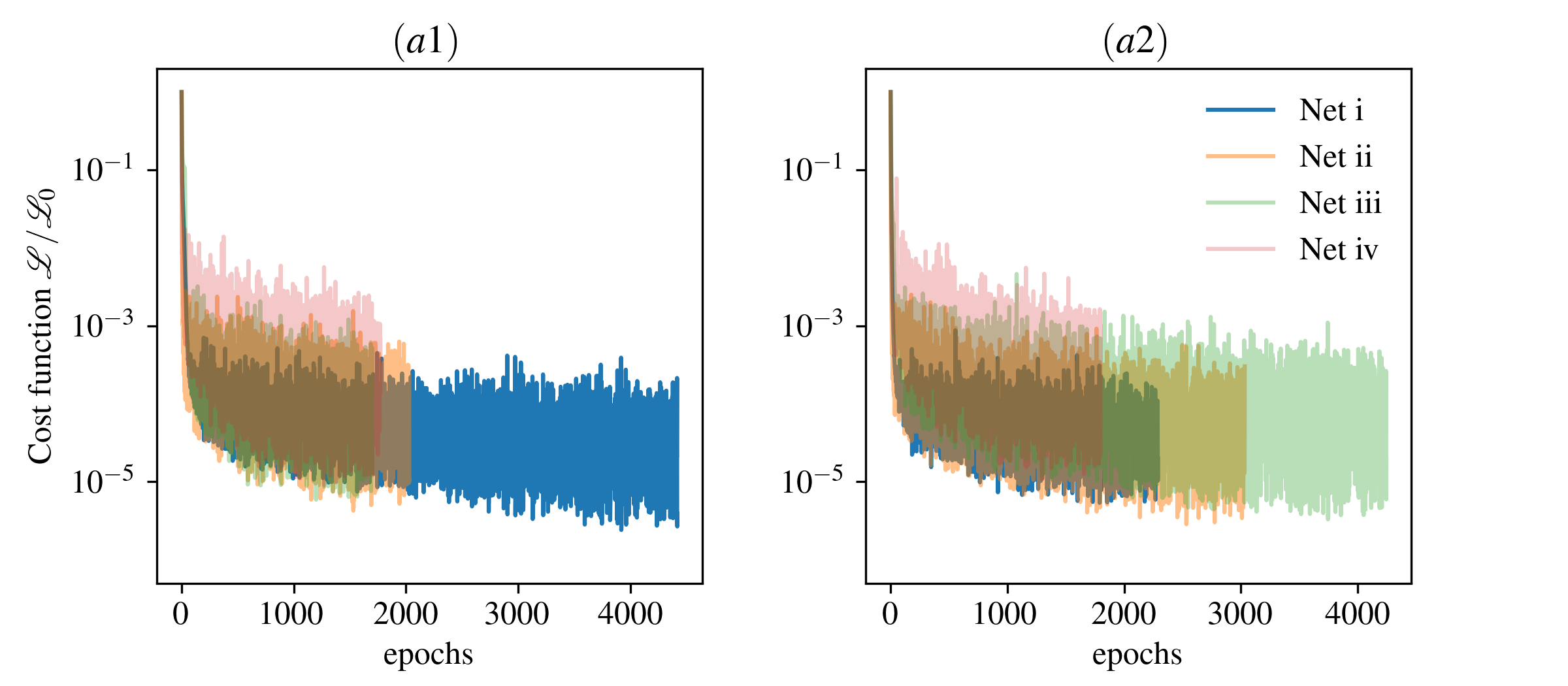}
    \caption{(a1)~Training on the IGA dataset using $u_x$, $u_y$, $\sigma_{xx}$, $\sigma_{yy}$, $\sigma_{xy}$, $f_x$ and $f_y$ components. (a2)~Learning with body forces $f_x$ and $f_y$ evaluated from centeral-differentiation of stress components.}
    \label{fig:pinn_iga_a1}
\end{figure}

Observing the lowest loss $\mathcal{L}$ on the analytical solution, we decided to study the influence of the global continuity of the numerical solution. We generated a $C^3$-continuous dataset using Isogeometric analysis \cite{bazilevs2010isogeometric}. We, therefore, analyze the system using $C^3$ IGA elements with again a grid of $40\times40$ dimension. The data are then mapped on to a grid of $100\times100$ and used to train the PINN models. The training results are shown in Fig.~\ref{fig:pinn_iga_a1}. The outputs are very similar to the high-order FEM datasets. 

\subsection{Identification using transfer learning}\label{sec:inversion}

Here we explore the applicability of our PINN framework to transfer learning: a neural network that is pre-trained is used to perform identification on a new dataset. The expectation is that since the initial state of neural network is not randomly chosen anymore, training should converge faster to the solution and parameters of the data. This is crucial for many practical aspects including adaptation to new data for online search or purchase history \cite{taylor2009transfer} or in geosciences, where we can train a representative PINN in highly-instrumented regions and use them at other locations with limited observational datasets. To this end, we use the pre-trained model on Net-iii (Fig.~\ref{fig:pinn_analytical_ab}), which was trained on a dataset with $\lambda=1.0$ and $\mu=0.5$ and then we explore how the loss evolves and the training converges when data is generated with different values of $\mu \in \left\{2.0, 1.5, 1.0, 0.1\right\}$. 

\begin{figure}[!ht]
    \centering
    \includegraphics[width=0.8\textwidth]{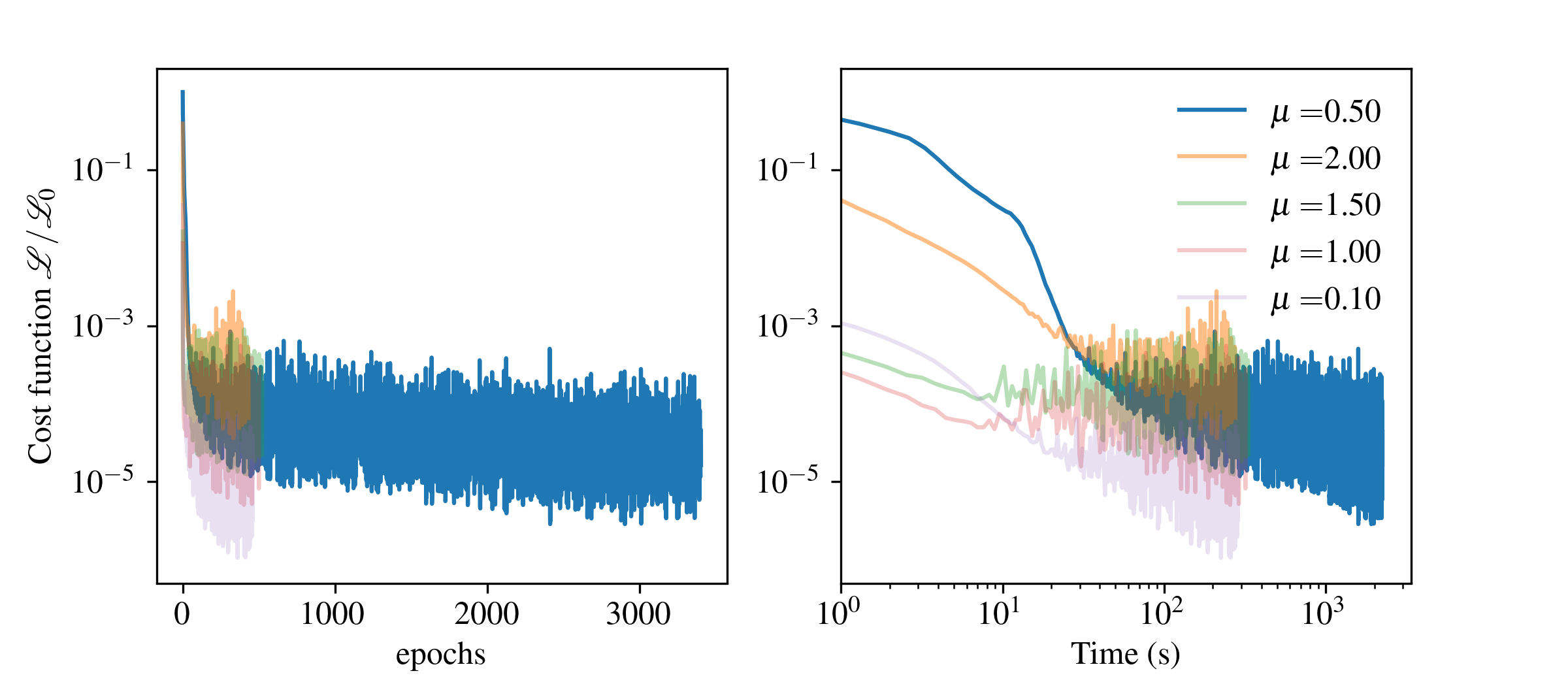}
    \caption{Identification of a new dataset generated with different values of $\mu$ using a pre-trained neural network on $\mu=0.5$. The re-training takes far less epochs to converge with an initial value for loss $\mathcal{L}$ much smaller.}
    \label{fig:pinn_identification}
\end{figure}

In Fig.~\ref{fig:pinn_identification} we show the convergence of the model with different datasets. Note that the loss is normalized by the initial value $\mathcal{L}_0$ from the pre-trained network on $\mu=0.5$ (Fig.~\ref{fig:pinn_analytical_ab}). As can be seen here, re-training on new datasets costs only a few hundred epochs with a smaller initial value for the loss. This is pointing to the advantage of deep learning and PINN, where retraining on similar data is much less costly than classical methods that rely on forward simulations. 

\subsection{Application to sensitivity analysis}\label{sec:sensitivity}

Performing sensitivity analysis is an expensive task when the analytical solution is not available, since it requires performing many forward numerical simulations. Alternatively, if we can construct a surrogate model to be a function of parameters of interest, then performing sensitivity analysis becomes tractable. However, construction of such a surrogate model is itself an expensive task within classical frameworks. Within PINN, however, this seems to be naturally possible. Let us suppose that the parameter of interest is shear modulus $\mu$. Consider an ANN model with inputs as $(x, y, \mu)$ and outputs as $(u_x, u_y, \sigma_{xx}, \sigma_{yy}, \sigma_{xy})$. We can, therefore, use a similar framework to construct a model that is a function of $\mu$ in addition to the space variables. Again, PINN can constrain the model to adapt to the physics of interest and therefore there is less data needed to construct such a model. 

\begin{figure}[!ht]
    \centering
    \includegraphics[width=.5\textwidth]{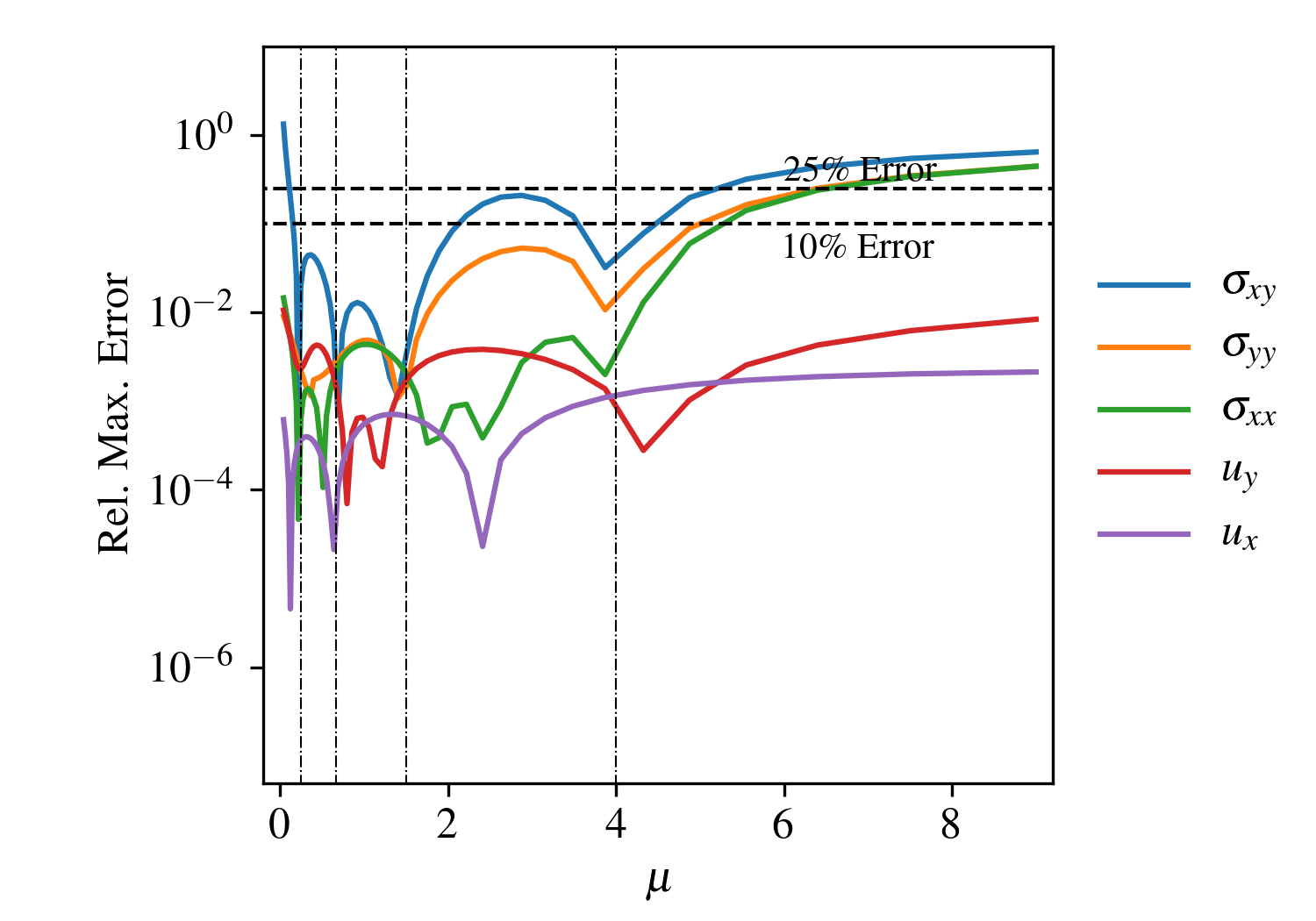}
    \caption{Application to sensitivity analysis: the model is trained on multiple datasets generated with different values of $\mu \in \left\{1/4, 2/3, 3/2, 4 \right\}$ (highlighted in dot-dashed lines). The model is then tested on a continuous range of values for $\mu \in (0, 9)$. The error is defined as $|\circ-\circ^*|/|\circ^*|$ at the point where $\circ^*$ is maximum. }
    \label{fig:pinn_sensitivity}
\end{figure}

Here, we explore if a PINN model trained on multiple datasets generated with various material parameters, i.e., different values of~$\mu$, can be used as a surrogate model to perform sensitivity analysis. The network in Fig.~\ref{fig:neural_network_arch} is now slightly adapted to carry $\mu$ as an extra input (in addition to $x, y$). The training set is prepared based on $\lambda=1$ and $\mu \in \left\{1/4, 2/3, 3/2, 4 \right\}$. Note that there is no identification in this case, and therefore the parameters $\lambda$ and $\mu$ are known at any given training data. The results of the analysis are shown in Fig.~\ref{fig:pinn_sensitivity}. For a wide range of values of $\mu \in (0, 9)$, the model performs very well in terms of displacements; it is less accurate, but still very useful, in terms of stresses with a maximum error for near-incompressible conditions, $\mu\approx0$.

\section{Extension to Nonlinear Elastoplasticity}

In this section, we discuss the application of PINN to nonlinear solid mechanics problems undergoing elastic and plastic deformation. We use the von~Mises elastoplastic constitutive model---a commonly used model to describe mechanical behavior of solid materials, in particular metals. We first describe the extension of the linear-elasticity relations in Eq.~\eqref{eqs:elas1} to the von~Mises elastoplastic relations. We then discuss the neural-network setup and apply the PINN framework to identify parameters of a classic problem: a perforated strip subjected to uniaxial extension.

\subsection{von~Mises elastoplasticity}

We adopt the classic elastoplasticity postulate of additive decomposition of the strain tensor~\cite{simohughes-ci},
\begin{equation}
    \varepsilon_{ij} = \varepsilon_{ij}^e + \varepsilon_{ij}^p.
\end{equation}
The stress tensor is now linearly dependent on the \emph{elastic} strain tensor:
\begin{equation}
    \sigma_{ij} = \lambda \varepsilon_{kk}^e \delta_{ij} + 2\mu\varepsilon_{ij}^e.
\end{equation}
The plastic part of deformation tensor is evaluated through a plasticity model. The von~Mises model implies that the plastic deformation occurs in the direction of normal to a yield surface $\mathcal{F}$ defined as $\mathcal{F}(\sigma_{ij}) := q - \sigma_Y$, as
\begin{equation}\label{eq:vonmises0}
    \varepsilon^p_{ij} = \gamma \frac{\partial \mathcal{F}}{\partial \sigma_{ij}},
\end{equation}
where~$\sigma_Y$ is the yield stress, $q$~is the equivalent stress defined as $q=\sqrt{3/2 s_{ij} s_{ij}}$, with $s_{ij}$ the components of the deviatoric stress tensor, $s_{ij} = \sigma_{ij} - \sigma_{kk}/3 \delta_{ij}$.  The strain remains strictly elastic as long as the state of stress~$\sigma_{ij}$ remains inside the yield surface, $\mathcal{F}<0$. Plastic deformation occurs when the state of stress is on the yield surface, $\mathcal{F}=0$. The condition $\mathcal{F}>0$ is associated with an inadmissible state of stress. Parameter~$\gamma$ is the plastic multiplier, subject to the condition $\gamma \ge 0$, and evaluated through a predictor--corrector algorithm by imposing the condition $\mathcal{F}\le 0$~\cite{simohughes-ci}. In the case of von~Mises plasticity, the volumetric plastic deformation is zero, $\varepsilon_{kk}^p = 0$. It can be shown that the plastic multiplier~$\gamma$ is equal to the equivalent plastic strain $\bar{\varepsilon}^p = \sqrt{2/3 e_{ij}^p e_{ij}^p}$, where $e_{ij}$ are the components of deviatoric strain tensor, $e_{ij} = \varepsilon_{ij} - \varepsilon_{kk}/3 \delta_{ij}$. 

Therefore, the elastoplastic relations for a plane-strain problem can be summarized as: 
\begin{equation}\label{eq:vonmises1}
\begin{split}
    \sigma_{ij, j} + f_i &= 0, \\
    \sigma_{ij} &= \lambda \varepsilon_{kk} \delta_{ij} + 2\mu\varepsilon_{ij} = (\lambda + 2/3\mu)\varepsilon_{kk} + s_{ij}, \\
    s_{ij} &= 2\mu (e_{ij}- e^p_{ij}), \\
    e^p_{ij} &= \bar{\varepsilon}^p \frac{\partial\mathcal{F}}{\partial\sigma_{ij}} = \bar{\varepsilon}^p \frac{2}{3}\frac{s_{ij}}{q}, \\
\end{split}
\end{equation}
subject to the elastoplasticity conditions:
\begin{equation}\label{eq:vonmises2}
    \bar{\varepsilon}^p \ge 0, \quad \mathcal{F} \le 0, \quad \bar{\varepsilon}^p \mathcal{F} = 0,
\end{equation}
also known as Karush--Kuhn--Tucker (KKT) conditions. For the von~Mises model, the plastic multiplier $\bar{\varepsilon}^p$ can be expressed as 
\begin{equation}\label{eq:vonmises3}
    \bar{\varepsilon}^p = \bar{\varepsilon}  - \frac{\sigma_Y}{3\mu} \ge 0,
\end{equation}
where $\bar{\varepsilon}$ is the total equivalent strain, i.e., $\bar{\varepsilon}=\sqrt{2/3e_{ij}e_{ij}}$. Therefore, the parameters of this model are the Lam\'e elastic parameters $\lambda$ and $\mu$, and the yield stress~$\sigma_Y$.

\subsection{Neural Network setup}
The solution variables for a two-dimensional problem are $u_x$, $u_y$, $\varepsilon_{xx}$, $\varepsilon_{yy}$, $\varepsilon_{xy}$, $\varepsilon_{xx}^p$, $\varepsilon_{yy}^p$, $\varepsilon_{zz}^p$, $\varepsilon_{xy}^p$, $\sigma_{xx}$, $\sigma_{yy}$, $\sigma_{zz}$, $\sigma_{xy}$. Since the out-of-plane components are no longer zero, they must be reflected in the choice of independent networks. Following the discussions for the linear elasticity case, we approximate the displacement and stress components $u_x, u_y, \sigma_{xx}, \sigma_{yy}, \sigma_{zz}, \sigma_{xy}$ with nonlinear neural networks as:
\begin{equation}\label{eq:vonmises4}
\begin{split}
    {u}_x &\approx \mathcal{N}_{u_x}(\mathbf{x}), \\
    {u}_y &\approx \mathcal{N}_{u_y}(\mathbf{x}), \\
    \sigma_{xx} &\approx \mathcal{N}_{\sigma_{xx}}(\mathbf{x}), \\
    \sigma_{yy} &\approx \mathcal{N}_{\sigma_{yy}}(\mathbf{x}), \\
    \sigma_{zz} &\approx \mathcal{N}_{\sigma_{zz}}(\mathbf{x}), \\
    \sigma_{xy} &\approx \mathcal{N}_{\sigma_{xy}}(\mathbf{x}). \\
\end{split}
\end{equation}
The associated cost function is then defined as
\begin{equation}\label{eqs:nn_vonmises}
\begin{split}
\mathcal{L} &= |u_x-u_x^*| + |u_y-u^*_y|  \\
            &+ |\sigma_{xx}-\sigma_{xx}^*| + |\sigma_{yy}-\sigma_{yy}^*| + |\sigma_{zz}-\sigma_{zz}^*| + |\sigma_{xy}-\sigma_{xy}^*| \\
            &+ |\sigma_{xx,x} + \sigma_{xy,y} - f_x^*| + |\sigma_{xy,x} + \sigma_{yy,y} - f_y^*|        \\
            &+ |(\lambda+2/3\mu)\varepsilon_{kk} + 2\mu (e_{xx}-e^p_{xx}) - \sigma_{xx}| \\
            &+ |(\lambda+2/3\mu)\varepsilon_{kk} + 2\mu (e_{yy}-e^p_{yy}) - \sigma_{yy}| \\
            &+ |(\lambda+2/3\mu)\varepsilon_{kk} + 2\mu (e_{zz}-e^p_{zz}) - \sigma_{zz}| \\
            &+ |2\mu (e_{xy}-e^p_{xy}) - \sigma_{xy}| + |(\bar{\varepsilon} - \sigma_Y/3\mu) - \bar{\varepsilon}^p| \\
            &+ |(1-\textrm{sign}(\bar{\varepsilon}^p))|\bar{\varepsilon}^p|| + |(1+\textrm{sign}(\mathcal{F}))|\mathcal{F}|| + |\bar{\varepsilon}^p\mathcal{F}| . \\
\end{split}
\end{equation}
The KKT positivity and negativity conditions are imposed through a penalty constraint in the loss function. For instance, $\bar{\varepsilon}^p \ge 0$ is incorporated in the loss as $(1-\textrm{sign}(\bar{\varepsilon}^p))|\bar{\varepsilon}^p|$. Therefore, for values of $\bar{\varepsilon}^p < 0$, the resulting `cost' is $2|\bar{\varepsilon}^p|$, which should vanish.

\subsection{Illustrative example}
\begin{figure}[H]
    \centering
    \includegraphics[width=\textwidth]{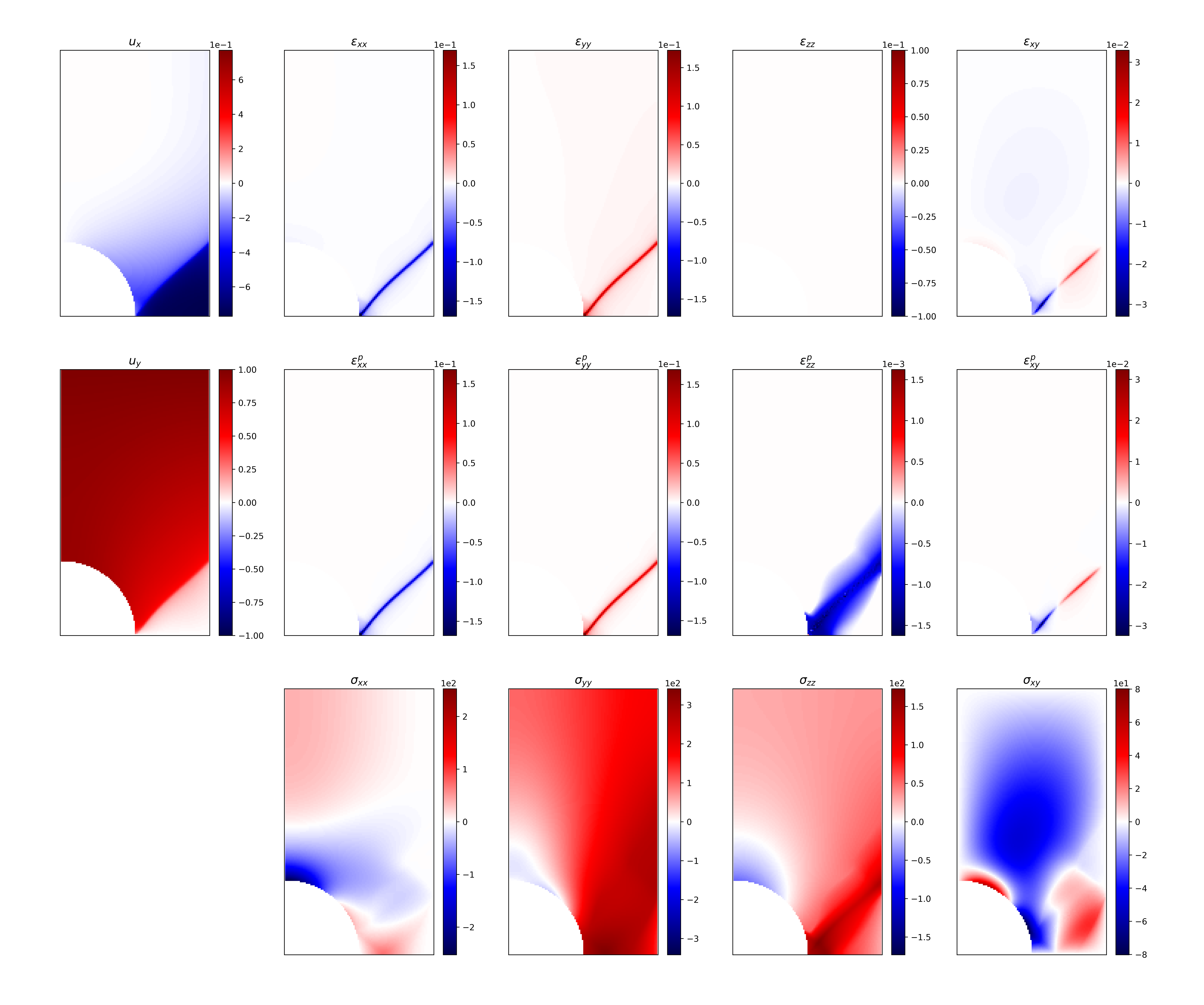}
    \caption{Reference solution of extension loading of a perforated plate from a high-fidelity FEM simulation. The true parameters are $\lambda=19.44~\textrm{GPa}$, $\mu=29.17~\textrm{GPa}$ and $\sigma_Y=243.0~\textrm{MPa}$. }
    \label{fig:vonmises-true}
\end{figure}

We use a classic example to illustrate our framework: a perforated strip subjected to uniaxial extension \cite{zienkiewicz1969elasto, simohughes-ci}. Consider a plate of dimensions $200~\text{mm}\times360~\text{mm}$, with a circular hole of diameter $100~\text{mm}$ located in the center of the plate. The plate is subjected to extension displacements of $\delta=1~\text{mm}$ along the short edge, under plane-strain condition, and without body forces, $f_i=0$. The parameters are $\lambda=19.44~\textrm{GPa}$, $\mu=29.17~\textrm{GPa}$ and $\sigma_Y=243.0~\textrm{MPa}$. Due to symmetry, only a quarter of the domain needs to be considered in the simulation. The synthetic data is generated from a high-fidelity FEM simulation using COMSOL software~\cite{COMSOL} on a mesh of 13041 quartic triangular elements (Fig.~\ref{fig:vonmises-true}). The plate undergoes significant plastic deformation around the circular hole, as can be seen from $\varepsilon_{ij}^p$ contours in~Fig.~\ref{fig:vonmises-true}. This results in localized deformation in the form of a shear band. While the strain exhibits localization, the stress field remains continuous and smooth---a behavior that is due to the choice of a perfect-plasticity model with no hardening.

\begin{figure}[H]
    \centering
    \includegraphics[width=\textwidth]{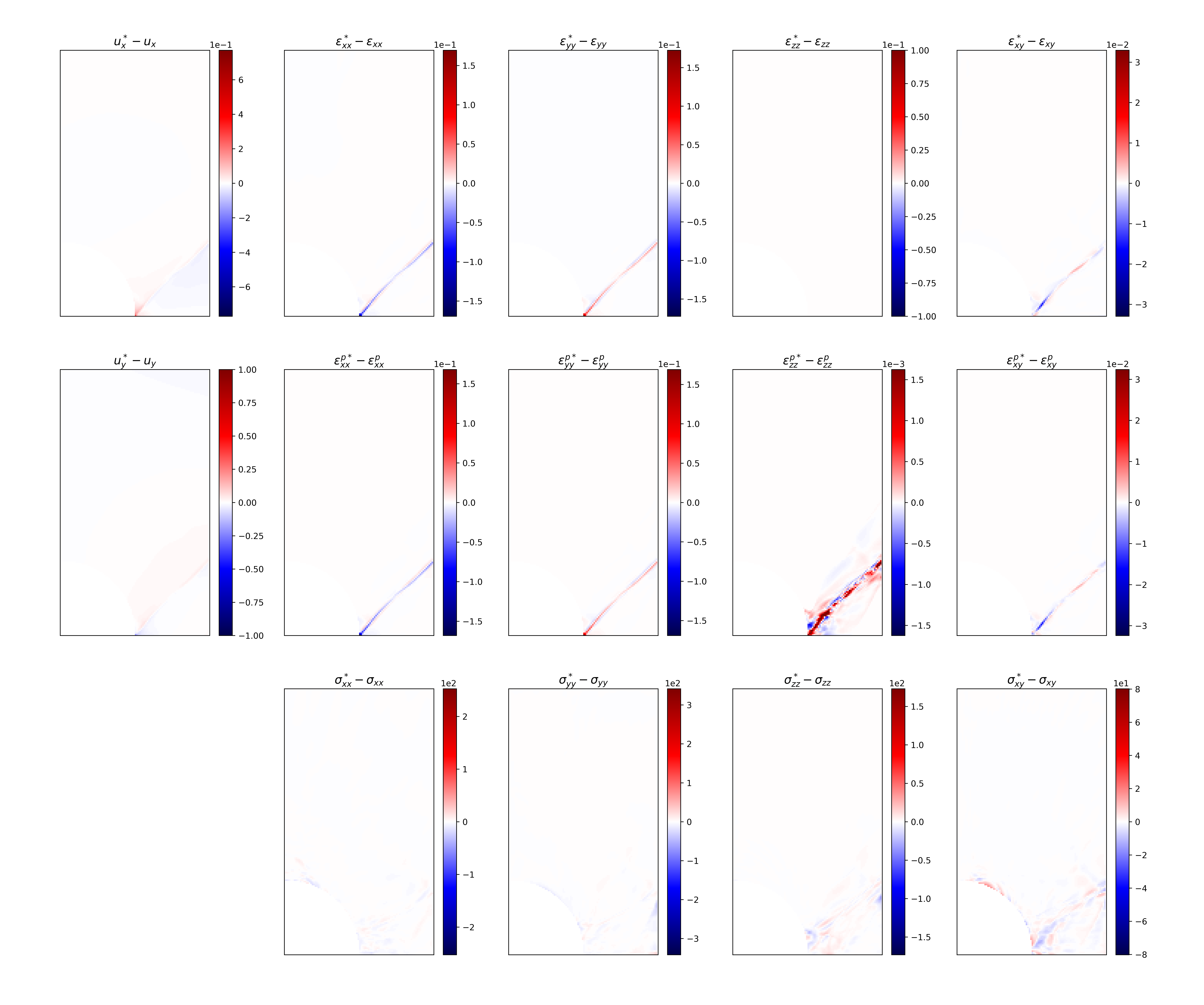}
    \caption{Error in predicted values from the PINN framework for displacements, strains, plastic strains and stresses. }
    \label{fig:vonmises-error}
\end{figure}

We use 2,000 data points from this reference solution, randomly distributed in the simulation domain, to provide the training data. The PINN training is performed using networks with 4~layers, each with 100~neurons, and with a hyperbolic-tangent activation function. The optimization parameters are the same as those used for the linear elasticity problem. The results predicted by the PINN approach match the reference results very closely, as evidenced by: (1)~the very small errors in each of the components of the solution, except for the out-of-plane plastic strain components (Fig.~\ref{fig:vonmises-error}); and (2)~the precise identification of yield stress $\sigma_Y$ and relatively accurate identification of elastic parameters $\lambda$ and $\mu$, yielding estimated values $\lambda=18.3~\textrm{GPa}$, $\mu=27.6~\textrm{GPa}$ and $\sigma_Y=243.0~\textrm{MPa}$. 

\section{Conclusions}
We study the application of a class of deep learning, known as Physics-Informed Neural Networks (PINN), for solution and discovery in solid mechanics. In this work, we formulate and apply the framework to a linear elastostatics problem, which we analyze in detail, but then illustrate the application of the method to nonlinear elastoplasticity. We study the sensitivity of the proposed framework to noise in data coming from different numerical techniques. We find that the optimizer performs much better on data from high-order classical finite elements, or with methods with enhanced continuity such as Isogeometric Analysis. We analyze the impact of the size and depth of the network, and the size of the dataset from uniform sampling of the numerical solution---an aspect that is important in practice given the cost of a dense monitoring network. We find that the proposed PINN approach is able to converge to the solution and identify the parameters quite efficiently with as little as 100 data points. 

We also explore transfer learning, that is, the use a pre-trained neural network to perform training on new datasets with different parameters. We find that training converges much faster when this is done. Lastly, we study the applicability of the model as a surrogate model for sensitivity analysis. To this end, we introduce shear modulus~$\mu$ as an input variable to the network. When training only on four values of $\mu$, we find that the network predicts the solution quite accurately on a wide range of values for $\mu$, a feature that is indicative of the robustness of the approach. 

Despite the success exhibited by the PINN approach, we have found that it faces challenges when dealing with problems with discontinuous solutions. The network architecture is less accurate on problems with localized high gradients as a result of discontinuities in the material properties or boundary conditions. We find that, in those cases, the results are artificially diffuse where they should be sharp. We speculate that the underlying reason for this behavior is the particular architecture of the network, where the input variables are only the spatial dimensions ($x$ and $y$), rendering the network unable to produce the required variability needed for gradient-based optimization that would capture solutions with high gradients. Addressing this extension is an exciting avenue for future work in machine-learning applications to solid mechanics.

\section*{Acknowledgements}
This work was funded by the KFUPM-MIT collaborative agreement `Multiscale Reservoir Science'.

\bibliographystyle{elsarticle-num-names} 
\bibliography{refs}

\end{document}